\documentclass[conference]{IEEEtran}
\IEEEoverridecommandlockouts
\pdfoutput=1
\usepackage{cite}
\usepackage{amsmath,amssymb,amsfonts}
\usepackage{algorithm}
\usepackage{algorithmic}
\usepackage{graphicx}
\usepackage{textcomp}
\usepackage{xcolor}
\def\BibTeX{{\rm B\kern-.05em{\sc i\kern-.025em b}\kern-.08em
		T\kern-.1667em\lower.7ex\hbox{E}\kern-.125emX}}

\usepackage{url}
\usepackage{subfigure}
\usepackage{amsthm}

\newtheorem{property}{Property}

\usepackage{verbatim}

\newcommand{\FUNCTION}[1]{\STATE \textbf{function} #1\textbf{:} \begin{ALC@g}}
	\newcommand{\ENDFUNCTION}{\end{ALC@g}}

\usepackage{amsmath,amsfonts,bm}

\def\eqref#1{equation~\ref{#1}}

\def\1{\bm{1}}

\DeclareMathAlphabet{\mathsfit}{\encodingdefault}{\sfdefault}{m}{sl}
\SetMathAlphabet{\mathsfit}{bold}{\encodingdefault}{\sfdefault}{bx}{n}

\usepackage{url}

\usepackage{graphicx}
\usepackage{amsmath}
\usepackage{amsthm}
\usepackage{amsfonts}
\newtheorem{problem}{Problem}
\usepackage{verbatim}
\usepackage{bm}
\usepackage{multirow}
\usepackage{subfigure}

\begin{document}
\title{FedCos: A Scene-adaptive Federated Optimization Enhancement for Performance Improvement} 


\author{\IEEEauthorblockN{Hao Zhang\IEEEauthorrefmark{1},
	Tingting Wu\IEEEauthorrefmark{1}, 
	Siyao Cheng\IEEEauthorrefmark{1}, and
	Jie Liu\IEEEauthorrefmark{2}}
	\IEEEauthorblockA{
		\IEEEauthorrefmark{1}Harbin Institute of Technology, Harbin, China\\
		\IEEEauthorrefmark{2}Harbin Institute of Technology (Shenzhen), Shenzhen, China\\
		Email: 
		zhh1000@hit.edu.cn,
		ttwu@ir.hit.edu.cn,
		\{csy, jieliu\}@hit.edu.cn}
}
\maketitle

\begin{abstract}

As an emerging technology, federated learning (FL) involves training machine learning models over distributed edge devices,  which attracts sustained attention and has been extensively studied. However, the heterogeneity of client data severely degrades the performance of FL compared with that in centralized training. It causes the locally trained models of clients to move in different directions. On the one hand, it slows down or even stalls the global updates, leading to inefficient communication. On the other hand, it enlarges the distances between local models, resulting in an aggregated global model with poor performance. Fortunately, these shortcomings can be mitigated by reducing the angle between the directions that local models move in. Based on this fact, we propose FedCos, which reduces the directional inconsistency of local models by introducing a cosine-similarity penalty. It promotes the local model iterations towards an auxiliary global direction.  Moreover, our approach is auto-adapt to various non-IID settings without an elaborate selection of hyperparameters. The experimental results show that FedCos outperforms the well-known baselines and can enhance them under a variety of FL scenes, including varying degrees of data heterogeneity, different number of participants, and cross-silo and cross-device settings. Besides, FedCos improves communication efficiency by 2 to 5 times. With the help of FedCos, multiple FL methods require significantly fewer communication rounds than before to obtain a model with comparable performance. 
\end{abstract}


\section{Introduction}

With the proliferation of sensing devices, a new era of Internet of Things (IoT) is sparked. These distributed devices generate significant amounts of data all the time, which promotes artificial intelligence into our life, such as smart healthcare system, automatic driving, smart city, etc. Traditionally, to reap the benefits of data of edge devices, the predominant approach is to collect all the data to the remote central cloud for processing and modeling. However, with the rapid development of IoT applications, transmitting the generated data results in high communication cost. Moreover, uploading the data may impose great privacy leakage risk. Under the limitation of legislation such as General Data Protection Regulation
(GDPR)~\cite{voigt2017eu},  training the deep learning model centrally by gathering data from users is impractical.

Edge computing is proposed to shift more computation to the network edge, allowing the edge devices to train models locally. However, insufficient data samples and local data shifts would lead to a worse model. 
With the landing of federated learning (FL), training deep learning models in parallel with the edge nodes becomes achievable.
FL is a distributed computing paradigm that multiple remote edge devices collaboratively train a global model without exchanging their local data. 
It treats the collaborative edge devices as working clients, training a machine learning model by local data and synchronizing the parameters via the parameter server. 
Since model parameters instead of raw data are transitted in the training process, the risk of privacy leakage is greatly reduced and the problem of communication overhead is alleviated\footnote{Generally, in IoT scenarios, the number of model parameters is relatively small compared to the massive amount of raw data continuously generated.}.

As an emerging machine-learning technique, FL is still in the early stages of research~\cite{wahab2021federated}. 
Compared with the traditional distributed machine learning, FL faces the challenge of data heterogeity caused by the limitation of data transferring.
Specifically, for the traditional distributed machine learning, where the training data of clients is sampled in IID (identically and independently distributed) way, the stochastic gradients of local models are unbiased estimates of the full gradients~\cite{rakhlin2012making, dean2012large}. In this case, all the clients have roughly the same optimal target. The local models move in the same direction. 
Therefore, the performance is almost identical with the centralized methods even if the local models are synchronized after multiple local iterations~\cite{lin2019don, yu2019parallel, ijcai2019-637}.
On the contrary, accuracy significantly degrades under more widespread non-IID data distributions~\cite{zhao2018federated, oza2021federated}. The heterogeneity of client data has been deemed as a pivotal factor suppressing the performance~\cite{chai2019towards} since the standard algorithm FedAvg~\cite{mcmahan2017communication} is proposed. 
In this situation, the condition of unbiased estimation is no longer met. 
Although it can be alleviated by reducing the number of local iterations (e.g., as an extreme case, all the local models are iterated only once in each communication round), intolerable communication overhead would be introduced. 
Therefore, \textit{how to enhance the learning performance with limited communication resources is a foundational goal of FL.}

Many previous works has been done to try to improve the performance of FL in a variety of aspects. 
Among them, plenty of studies focus on local training~\cite{li2018federated, yao2019federated,  wang2020tackling}. For instance, FedProx~\cite{li2018federated} adds a proximal term to restrict the local model not far from the current global one.
FedMMD~\cite{yao2019federated} has the same goal but by making their output distribution similar.
But they may slow down model updates since they enforce the local models close to the stale model. 
Besides, from the experimental study~\cite{li2021federated}, FedAvg still performs best in many kinds of FL scenes.
Other approaches attempt to improve aggregation scheme~\cite{chen2021fedbe, NEURIPS2020_18df51b9, al-shedivat2021federated, NEURIPS2020_ac450d10}, which either requires additional public data for model distillation~\cite{NEURIPS2020_18df51b9} or introduces expensive training costs for obtaining sufficient model samples~\cite{chen2021fedbe}, which is not suitable for edge devices.
Some works~\cite{karimireddy2020scaffold, acar2021federated} attempt to speed up training to reduce the communication rounds, but they have no effect on improving performance or even hurt performance~\cite{li2021federated}.
Thus, in practice FedAvg still is the widely accepted one.

In the non-IID data scenarios, the local model of each client updates iteratively to its local optimum based on the data itself. 
During the local iterations, local models move in diverse directions, which causes the two following problems. 
Firstly, the gains from local training would be offset by the aggregation of local models, 
which slows down or even stalls the global updates to \textit{lead to inefficient communication}.
Secondly, local models are far away from each other, 
which causes the aggregated model distant from all local models, \textit{leading to poor model performance} on all local data. What's worse, to reduce the communication cost, the clients usually perform multiple SGD steps before aggregation, which enlarges the distances further. 
From our investigation, these shortcomings can be addressed by reducing the angle between the directions that local models move in. Based on this fact,  we propose a new \textbf{Fed}erated enhancement with \textbf{Cos}ine-similarity penalty (FedCos). We introduce an auxiliary global direction that all clients refer to in the local training phase to reduce the directional inconsistency of local models. This constraint accelerates the global updates and diminishes the distances between local models so that the aggregated model is not far away from all the local optima. 
By analyzing the execution process, we observe FedCos explores more points around the convergence point of FedAvg in the parameter space, which facilitates more points closer to the global optimum to be found.\footnote{In fact, due to the nonconvexity of neural network, SGD and other optimization methods aim to find minima. Here we do not distinguish minima and optima, which is not affect the analysis of this paper.}  
Meanwhile, FedCos is auto-adapt to the FL settings. The effect of penalty is tuned automatically according to the degrees of heterogeneity of data. Elaborate hyperparameter selection for different scenes is no longer required.
In conclusion, the main contributions are summarized below:
\begin{itemize}
	\item 
	We investigate how data heterogeneity leads to inefficient communication and performance degradation in detail, and explore the angle between the directions that local models move in is the critical issue.
	\item 
	We propose FedCos, which reduces the directional inconsistency of local models by introducing a cosine-similarity penalty. FedCos can obtain better models than FedAvg and is auto-adapt to the settings without elaborate hyperparameters selection.
	\item 
	We construct a wide variety of FL scenes comprising different degrees of data heterogeneity, varying amounts of participants, under cross-silo and cross-device settings. FedCos outperforms other well-known FL methods (FedAvg, FedProx, FedOpt and FedAvgM) regardless of FL scenes, and can enhance them. Furthermore, FedCos improves the efficiency of communication. It greatly reduces the number of communication rounds to obtain global models with the same performance. To our best knowledge, FedCos is the first enhancement that can persistently outperform  and enhance FedAvg and other FL methods in a variety of scenes.
\end{itemize}



\section{Related work}\label{relatedwork}

As an extension of distributed training, McMahan et al.~\cite{mcmahan2017communication} first propose the concept of federated learning and related training paradigm FedAvg. 
After that, a lot of attention has been attracted to explore its potential and applicability.
It has been shown that the data heterogeneity of clients induce performance decline~\cite{zhao2018federated, haddadpour2019convergence, li2019convergence}. To address this issue, existing methods primarily adjust the local training. Among them, FedProx~\cite{li2018federated} adds a regularization to the local loss function, which enables the local parameters not far from the global parameters. FedMMD~\cite{yao2019federated} tries to constrain the distribution of the local model close to the global one. SCAFFOLD~\cite{karimireddy2020scaffold} manually modifies the drift of local training. 
However, most of them fail to outperform FedAvg in multiple FL scenarios.
Another approach improves the performance by modifying the aggregation scheme. For example, FedBE~\cite{chen2021fedbe} introduces a Bayesian scheme, where the global model is regarded as the expectation of model distribution, but too much training cost is introduced. FedDF~\cite{NEURIPS2020_18df51b9} introduces data distillation technology to distill the global model from multiple local models, but additional public data is needed. FedPA~\cite{al-shedivat2021federated} constructs a posterior model replacing the weighted average model. The prior assumption of uniform distribution is suspicious. DRFA~\cite{NEURIPS2020_ac450d10} aims to solve a more general problem with arbitrary weights for local clients, which makes the problem harder.

Additionally, some studies enhance the performance under specific context. GKT~\cite{NEURIPS2020_a1d4c20b} focuses on the scenario where the edge has limited resources. FedRobust~\cite{NEURIPS2020_f5e53608} considers the data distribution has a common drift on one client. CFL~\cite{9174890} and IFCA~\cite{NEURIPS2020_e32cc80b} divide the clients into several classes and learn a model for each class, where the data distribution is heterogeneous. These studies for specific problems are unsuitable for general FL scenes.
Besides accuracy, FL also faces other challenges~\cite{wahab2021federated} such as communication, privacy, security, personality, which also arouse wide attentions~\cite{liang2020think, li2021ditto, yu2020fed+, diao2021heterofl, ozdayi2020defending} but are out of the scope of this paper. In addition, many traditional studies such as the Ascent Method~\cite{shalev2013stochastic} and ADMM~\cite{boyd2011distributed} focus on similar problems with theoretical guarantees, which are not generalizable to the deep learning-based FL, where strong duality is no longer satisfied. 

\begin{figure}[htbp]
	\centering
	\includegraphics[width = 1\columnwidth]{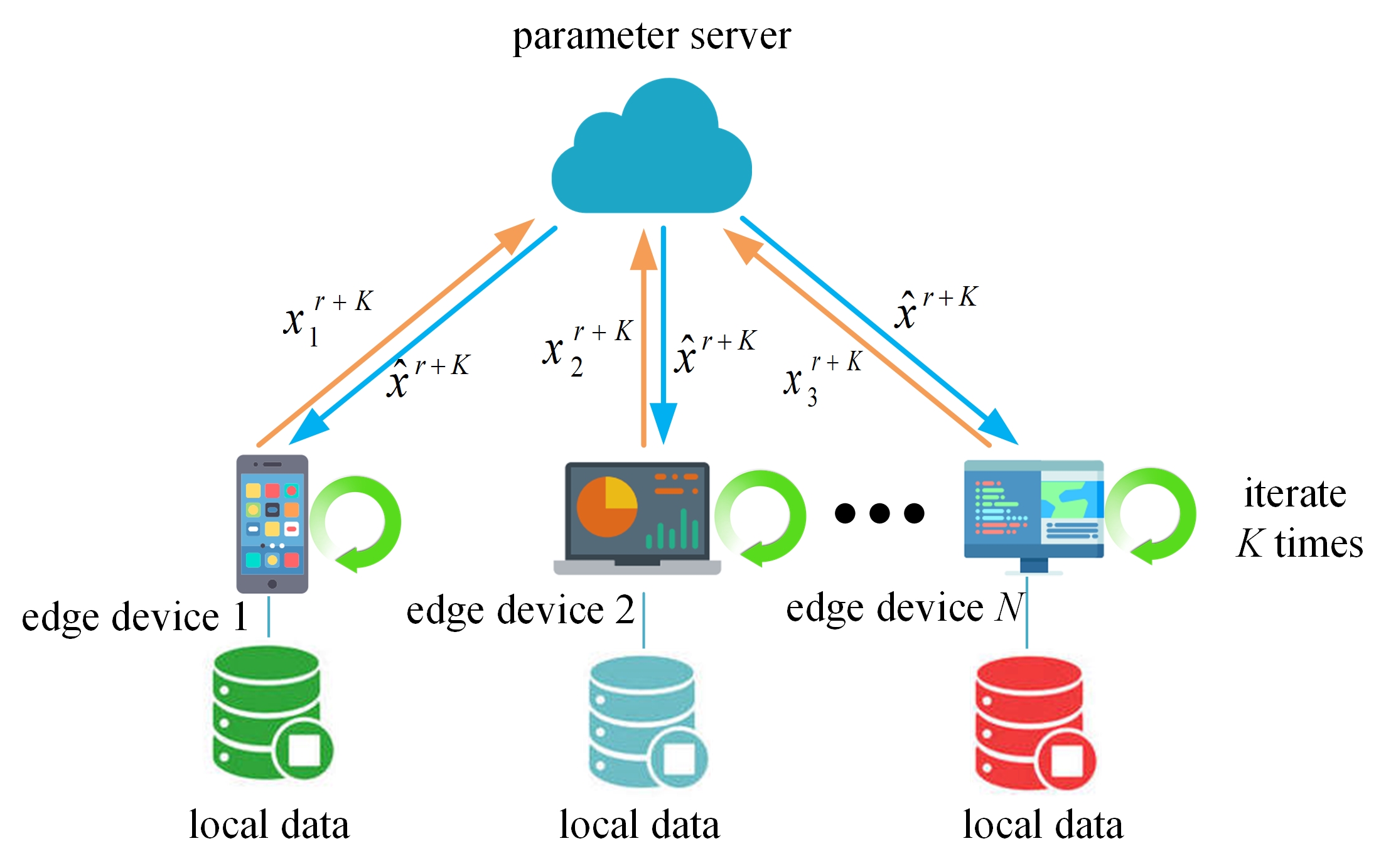} 
	\caption{The federated learning structure for distributed edge data.}
	\label{fig:fl} 
\end{figure}

\section{Problem statement}\label{limitations}

\subsection{Background: FedAvg structure}
In this paper, we discuss the distributed network as shown in Fig.~\ref{fig:fl}. It involves $N$ edge devices as clients and one parameter server to jointly learn a global model $x \in \mathbb{R}^m$ without data sharing by federated learning (FL). Each device $i \in [1,N]$ owns one local dataset $\mathcal{D}_i$. 
The aim can be formulated as the following problem
\small
\begin{equation}\label{eq:fl}
	\arg\min_x {f(x) = \sum_{i=1}^N \lambda_i f_i(x)},
\end{equation}
\normalsize
where 
$f_i$ represents the loss function on client $i$, and $\lambda_i$ is the weight satisfying $\sum_{i=1}^N \lambda_i=1$. 
To facilitate this, Federated Averaging algorithm (FedAvg) is proposed in the federated setting, which tackles the problem iteratively. More specifically, each client performs SGD up to $T$ steps and synchronizes the model every $K$ steps, i.e., synchronization happens at steps $\mathcal{I}=\{nK|n=1,2,\cdots\}$. There are two phases in each federated learning round: 

	\textbf{Local training (on client).} For client $i \in \mathcal{S}$, where $\mathcal{S}$ denotes the set of clients participating in the training. The local model iterates from $x_{i}^{r}$ ($r \in \mathcal{I}$), which is initialized to the latest received global model $\hat{x}^{r}$, and updates as 
	\small
	\begin{equation}\label{eq:iteration}
		x_{i}^{r+t+1} = x_i^{r+t} - \eta_i \nabla f_i(x_i^{r+t}, B_i), 
	\end{equation}
	\normalsize
	where $t=\{0, \cdots, K-1\}$, $B_i$ is the mini-batch sampled from client $i$'s local data $\mathcal{D}_i$, and $\eta_i$ is the learning rate. After $K$ steps, the client sends the  local model $x_{i}^{r+K}$ to the server.
	
	\textbf{Global aggregation (on server).} The global model is updated as follows:
	\small
	\begin{equation}\label{aggregation}
		\hat{x}^{r+K} = \sum_{i\in \mathcal{S}} \lambda_i x_{i}^{r+K},
	\end{equation}
	\normalsize
	where $\lambda_i = \frac{|\mathcal{D}_i|}{\sum_{j \in \mathcal{S}} |\mathcal{D}_j|}$, and then the global model parameters are distributed to all or selected clients for the next ``local training and global aggregation'' round.

In each round, some or all clients participate in the training, which corresponds to distinct FL scenarios called ``cross-device'' or ``cross-silo'' respectively.

\subsection{Directional inconsistency}\label{limations_of_fedavg}

Data heterogeneity of clients, i.e., non-IID data setting, is a common phenomenon in FL, which seriously affects FedAvg's performance. 
Although numerous methods are proposed, few approaches exceed FedAvg in all scenes.  
Li et al~\cite{li2021federated} exhaustively compare FedAvg and other three typical improved methods (FedProx~\cite{li2018federated}, SCAFFOLD~\cite{karimireddy2020scaffold} and FedNova~\cite{wang2020tackling}), finding out 
FedAvg still exceeds others in a variety of non-IID scenarios.  
As the standard solution of FL, FedAvg is still the most appropriate baseline that most studies concern.

In non-IID scenarios, the local model of each client updates iteratively to its local optimum.  
During the local iterations, local models move in different directions, which causes two  problems as follows:
\begin{problem}[Communication inefficiency]
	Heterogeneity slows down, even stalls the global updates, making the training inefficient, even aborting the training halfway in practice.
\end{problem}

In each round, we define the displacement vector of client $i$ after the local training phase as $d_i = x_i^{r+K} - \hat{x}^r$. 
Based on 
\eqref{aggregation}, we obtain the displacement vector of the global model $\hat{d} = \sum_{i=1}^N \lambda_i d_i$ whose norm representing the moving distance of the global model can be denoted as
\small
\begin{equation}\label{hatd}
	\|\hat{d} \| = \|\sum_{i=1}^N \lambda_i d_i\| = \sqrt{ \sum_{i=1}^N \lambda_i^2 \|d_i\|^2 + \sum_{i \neq j} \lambda_i \lambda_j \|d_i\| \|d_j\| \cos \theta_{ij}},
\end{equation}
\normalsize
where $\theta_{ij}$ is the angle between two vectors $d_i$ and $d_j$.
When we fix $\|d_i\|$ and $\lambda_i$ ($1\leq i \leq N$) (which are dominated by the learning rate and data amount respectively), $\|\hat{d} \|$ is determined by $\theta_{ij}$.

\begin{figure}
	\centering
	\includegraphics[width = 1\columnwidth]{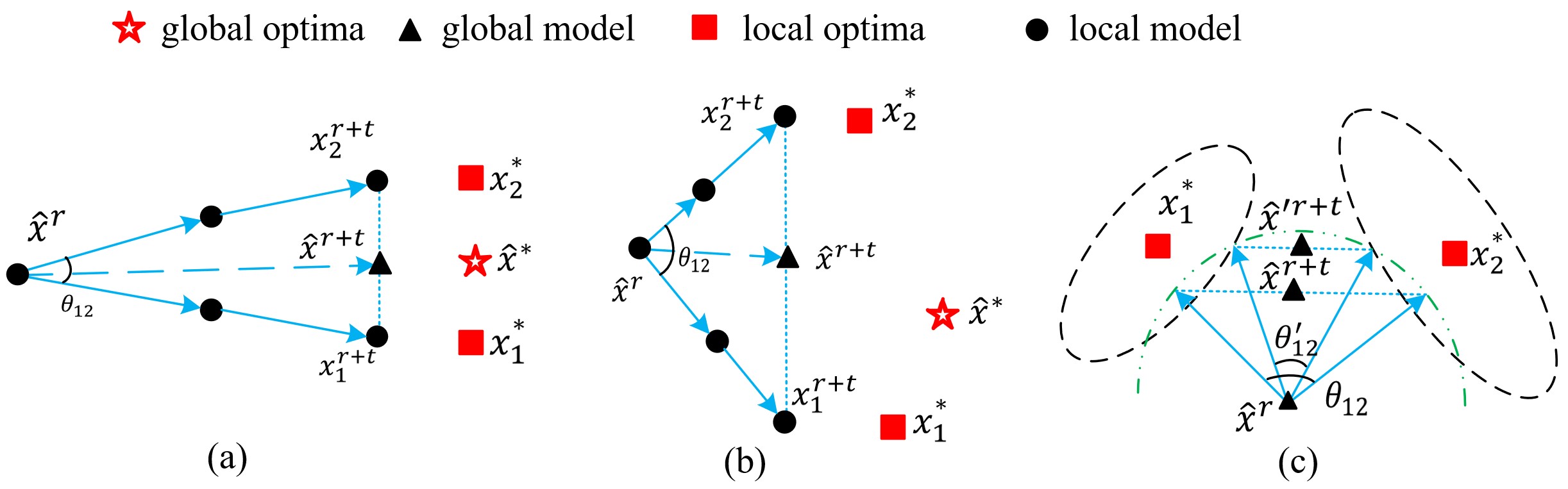} 
	\caption{Examples of FL update with two clients.(a) IID setting; (b) Non-IID setting;(c) adjusting the angle between the directions that local models move in.}
	\label{fig:fl_move} 
\end{figure}

If the data distribution among the parties is homogeneous (i.e., IID), for any $x$, $i$ and $j$, $\nabla f_i(x) \approx \nabla f_j(x)$. As a result, all the local models move roughly in the same destination (Fig.~\ref{fig:fl_move}(a)), namely $\theta_{ij} \approx 0$. If $\|d_i\|=\|d\|$ ($1\leq i \leq N$), we have $\|\hat{d} \| = \| d\|$. Thus in IID setting, the moving distance of the global model is as large as the local one.

\begin{figure}[tbp]
	\centering
	\subfigure[The cosine of the angle between two local models in non-IID scenario]{
		\includegraphics[width=0.7\columnwidth]{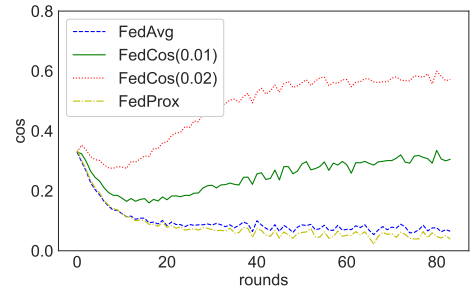} 
		\label{fig:consine_noniid3} 
	}
	\subfigure[The accuracy on test dataset of global models]{
		\includegraphics[width=0.7\columnwidth]{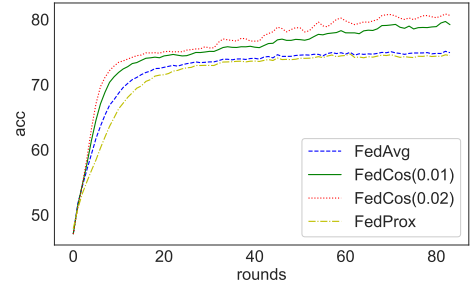} 
		\label{fig:consine_noniid5} 
	}
	\caption{Reducing the inconsistency of directions of clients' models helps improve model performance. The experiment settings are the same as those in Fig.~\ref{fig:fedcospro}.}
	\label{fig:consistency} 
\end{figure}

If the data distribution among the clients is heterogeneous (i.e., non-IID), $\nabla f_i(x)$ and $\nabla f_j(x)$ are quite different. 
In this case, 
all the local models move in different direction (Fig.~\ref{fig:fl_move}(b)). Since in high-dimensional space, any two vectors are almost perpendicular, i.e., $\theta_{ij} \approx 90^{\circ}$. If $\|d_i\| =\|d\|$ and $\lambda_i=\lambda$ ($1\leq i \leq N$), we have $\|\hat{d} \| = \frac{\|d\|}{\sqrt{N}}$, 
which means in the non-IID data settings, the global model moves far less distance than local one. When $N$ is large, the global model updates are very slow, even stalled ($N \to \infty$). To mitigate this problem, from 
(\ref{hatd}), a smaller $\theta_{ij}$ leads to a larger $\|\hat{d} \|$, which means the global model moves further. It also be verified in Fig.~\ref{fig:consine_noniid3} and Fig.~\ref{fig:fedcospro1} in the experiment section.
\begin{problem}[Poor performance]
	The convergence point of FedAvg is far from the true global optimum in non-IID settings.
\end{problem}
Based on Taylor's theorem, if the distance between a point $x$ and the optimum of one client (e.g., $x^*_1$) is small, $x$ is a suitable model for its local data. Therefore, we aim to find a point close to the optima of all clients. Under the non-IID condition, the distances between the local optima of different clients are large. If $x$ is too close to the optimum of one client, it usually has bad performance for other clients. The global optimum achieves the best balance between different clients, which is not too far away from all the local optima. 
Actually, 
Xie et al.~\cite{xie2021a} prove that SGD favors the flat minima. Izmailov et al.~\cite{izmailov2018averaging} also point out that the optimum is located in a flat region, where the performance is almost the same (referred to as ``optimal region'' for short). SGD usually stays at a point at the edge of the region. Hence it is feasible to find a representative point on each optimal region, which is closer to the optimal regions of other clients. 
As shown in Fig.~\ref{fig:fl_move}(c), if we adjust the angles between the directions that local models move in (e.g., $\theta_{12} \to \theta_{12}'$), and keep the local models in or at the edges of the optimal regions, the new aggregated model (e.g., $\hat{x}'^{r+t}$) is better than the previous one (e.g., $\hat{x}^{r+t}$) for being closer to the local models. Suppose there are two clients satisfying  $\|d_1\| = \|d_2\| = \|d\|$ and $\lambda_1 =\lambda_2$, and the angle between $\|d_1\|$ and $\|d_2\|$ is $\theta_{12}$. The distance between global model and local model is $\| d \|\cos\frac{\theta_{12}}{2}$. A smaller angle between $d_1$ and $d_2$ may lead to a better aggregated model. It also be verified in Fig.~\ref{fig:consine_noniid3} and Fig.~\ref{fig:fedcospro4} in the experiment section.

From the above investigation, reducing the directional inconsistency of local models
would be a critical issue to mitigate the influence of data heterogeneity, which motivates us to introduce a cosine-similarity penalty to reduce the angle between the directions that local models move in.



\begin{algorithm}[htbp]
	\caption{\textbf{FedCos}: \textbf{Fed}erated learning with \textbf{Cos}ine-similarity penalty} 
	\label{al:FedCos}
	\begin{algorithmic}[1]
		\REQUIRE 
		\hspace*{0.02in} $K$, $T$, $\hat{x}^0$, $\hat{d}^0=\vec{0}$ \\
		
		\STATE Server sends $\hat{x}^0$ and $\hat{d}^0$ to all clients.
		\FOR {$t = 1$ to $T$}
		\STATE (Local training:)
		\STATE Each client in $\mathcal{S}$ updates the local model $x_i^t$ by minimizing 
		(\ref{eq:fedcos}).
		\IF {$t\%K == 0$} 
		\STATE Each client in $\mathcal{S}$ sends $x_i^t$ to the server.
		\STATE (Global aggregation:)
		\STATE The server updates $\hat{x}^t$ and $\hat{d}^t$ by 
		(\ref{aggregation}) and 
		(\ref{eq:direction}).
		\STATE The server randomly selects a subset of clients $\mathcal{S}$, and sends $\hat{x}^t$ and $\hat{d}^t$ to the clients in $\mathcal{S}$.
		\ENDIF 
		\ENDFOR
	\end{algorithmic}
\end{algorithm}

\section{Method}\label{ourmethod}

\subsection{FedCos}\label{FedCos}

Based on the above analysis, we try to decrease the angle between the directions that any two local models move to in the local training phase. However, it is impractical to get others' direction vectors. Therefore, we construct a global direction to constrain the local model updated along this direction. By adding this auxiliary direction, all the local models update towards the same direction to mitigate the direction inconsistency. Meanwhile, we use the auxiliary direction in which the global model moves, the performance degradation of global model is avoided. Based on this design, the loss function of the local model is
\small 
\begin{equation}\label{eq:fedcos}
	L_i(x_i; \hat{x}^r, \hat{d}^r) = f(x_i) + \mu_i (1-\cos\theta_i),  
\end{equation} 
\normalsize
where $\mu_i$ is the weight of directional consistency, $\hat{x}^r$, $\hat{d}^r$ are the initial global model and the global direction in this round, and $\theta_i$ is the angle between $(x-\hat{x}^r)$ and $d^r$, so 
\small
\begin{equation} 
	\cos\theta_i = \frac{\langle x_i-\hat{x}^r, \hat{d}^r\rangle}{\|x_i-\hat{x}^r \| \cdot \| \hat{d}^r\|}.
\end{equation} 
\normalsize
We define $\theta_i = 0$ (i.e., $\cos\theta_i =1$) if $\|x_i-\hat{x}^r \|$ or $\| \hat{d}^r\|$ is equal to $0$.
We employ the direction in which the latest global model is moving as the reasonable global direction, 
i.e., 
\small
\begin{equation}\label{eq:direction}
	\hat{d}^r = \hat{x}^r - \hat{x}^{r-K},
\end{equation}
\normalsize
where $\hat{x}^{r-K}$ is the global model of the last round. Base on the above formation, we propose our algorithm FedCos (\textbf{Fed}erated learning with \textbf{Cos}ine-similarity penalty) and summarize it in Algorithm~\ref{al:FedCos}. FedCos is similar to FedAvg with a few differences. In the local training phase, instead of just minimizing the loss $f_i(x)$ based on the local data, each client adds a direction penalty. In the global aggregation phase, the server also calculates the new global direction referenced in the next round besides aggregation as FedAvg.

Moreover, we do not limit the form of $f_i(x_i)$ in 
(\ref{eq:fedcos}). Therefore, \emph{FedCos also can be regarded as an increasement on other methods of federated learning}. For instance, if $f_i(x_i)$ is the cross entropy function, the basic method is FedAvg. If $f_i(x_i)$ is the cross entropy function adding a term $\| x_i - \hat{x}^r\|^2$, the basic method is FedProx.

\subsection{Property analysis}\label{property}

We discuss some advantages of FedCos.

\begin{property}\label{pro:consistency}
	FedCos can mitigate the inconsistency of directions of clients' models. 
\end{property}

This property is the motivation of FedCos, which is consistent with the practical experiments. In Fig.~\ref{fig:consistency}, we illustrate the dirctional inconsistency by randomly selecting two local models (indexed by $i$ and $j$). In Fig.~\ref{fig:consine_noniid3} shows the change of angle of the two local models (i.e., the angle between $x_i^{r+K} - \hat{x}^r$ and $x_j^{r+K} - \hat{x}^r$) in  non-IID scenario at each aggregation. For FedAvg, as the training progresses, the direction of local clients become different. 
By contrast, FedCos reduces the inconsistency with smaller angles. 
Fig.~\ref{fig:consine_noniid5} illustrates the performances on test dataset for  relevant models. It shows that FedCos with less inconsistency outperforms other two methods. Therefore, reducing the inconsistency of directions of clients' models helps improve model performance.

\begin{property}
	FedCos is auto-adapt to data heterogenity without selecting penalty weight elaborately.
\end{property}
For FedCos, the update of local model is controlled by $f_i(x)$ and $h(x) = \mu_i (1-\cos\theta_i)$ 
(\ref{eq:fedcos}), where $0 \leq h(x) \leq 2\mu_i$ and usually $f_i(x) \geq 0$. 
The influence of penalty veries for different data heterogenity. 
For IID scenarios, all the local models of clients move in the same direction. The given global direction is consistent with the local optimal ones ($\cos \theta_i \approx 1$). In this situation, FedCos automatically degrades to FedAvg when the penalty term approaches 0.
For non-IID scenarios, more serious data heterogeneity leads to more direction inconsistency of local model updates, which enhances the effect of penalty even for a fixed $\mu_i$ (i.e., a bigger $h(x)$), and vice versa. 

In the local training phase, $f_(x)$ is dominant and $h(x)$ has little influence in the first few iterations (usually we choose a small $\mu_i$). FedCos has little difference from FedAvg. When $f_i(x)$ approaches $0$, i.e., the local model is close to the local minimum, $h(x)$ becomes more effective. 
This penalty only constrains the orientation of local model updates.
Since there is a flattened optimal region~\cite{izmailov2018averaging, xie2021a}, direction variation has little influence on accuracy once the model reaches the area, which guarantees $f_i(x)$ and $h(x)$ get small simultaneously. Therefore, FedCos cannot has little influence on the performance of local model. Furthermore, due to reducing the model inconsistencies by the same auxiliary direction, an better aggregated model can be gained.

Therefore, FedCos provides a universal performance improvement scheme, which is auto-adapt to the changes of data heterogeneity. 
We do not need to elaborately select the penalty weight regardless of the non-IID degrees of data. A fixed one would perform well for many kinds of FL settings, which is verified in the following experiments.



In contrast, FedProx reduces the differences between the local models and the aggregated global model by adding the penalty $\| x - \hat{x}\|^2$ with weight $\mu$, which plays an invariant role no matter how heterogeneous the data is. Moreover, for IID scenarios, there is little difference between the local models, but the penalty takes an opposite effect that prevents the model from approaching the optimum. Thus FedProx has to develop a complex heuristic method to tune $\mu$ carefully (section 5.3 in~\cite{li2018federated}).

\begin{figure}[tbp]
	\centering
	\includegraphics[width = 1\columnwidth]{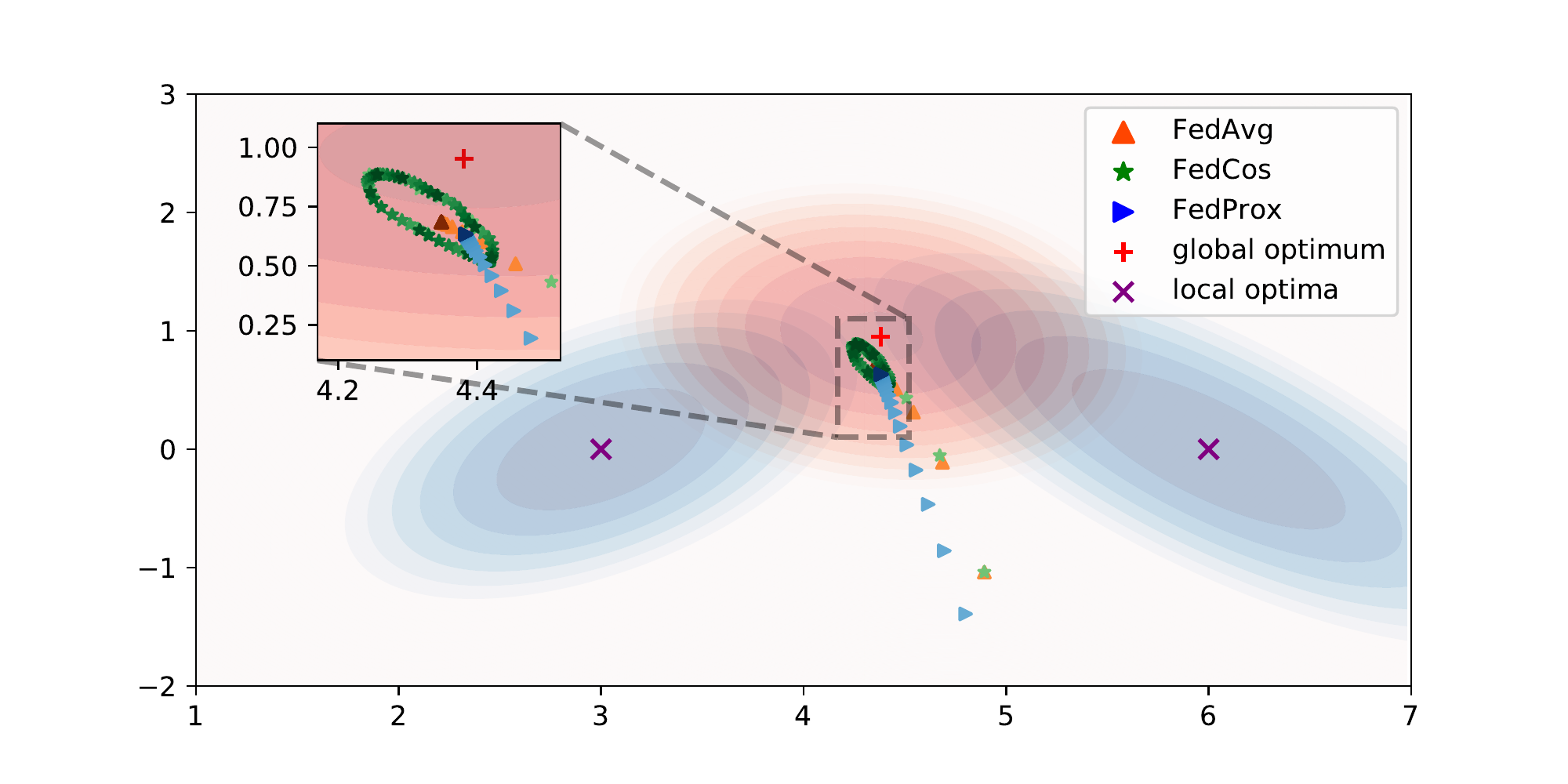}  
	\caption{Comparing the global model trajectories among FedAvg, FedProx and FedCos with 80 rounds. The points with darker colors denote the models from later rounds.}
	\label{fig:example_com} 
\end{figure}

\section{performance analysis}

To illustrate the properties of FedCos, we construct a simple two-dimensional federated example Fig.~\ref{fig:example_com}. 
There are two clients with local convex optimization problems denoted by $f_1(x)$ and $f_2(x)$.\footnote{In the example, the explicit functions are $f_1((x_{(0)}, x_{(1)})) = 0.5(x_{(0)}-6)^2+0.75 (x_{(0)}-6)x_{(1)}+0.5x_{(1)}^2$, $f_2((x_{(0)}, x_{(1)})) = 0.5(x_{(0)}-3)^2-0.5 (x_{(0)}-3)x_{(1)}+0.5x_{(1)}^2$.
The initial point of model is $(5.1, -3.1)$.} The global optimization problem is $f(x) = f_1(x)+f_2(x)$. Since $f_1(x)$ and $f_2(x)$ are quite different (which simulates the non-IID scenario), the optimum of $f(x)$ is different with both local ones. With the same parameter settings (such as the learning rate, the number of local steps, and the number of rounds), we run FedCos, FedAvg and FedProx respectively. For FedAvg, it cannot achieve the global optimum. 
Specifically, the local models (almost) reach the local optima after iterating enough steps at the end of one round. In this case, the global model is close to the middle point of two local optima, rather than the true global optimum. FedProx, which attempts to deal with this problem by decreasing the distances of local models, forces the local models updating near the global model of the last round. 
Unfortunately, FedProx fails to outperform FedAvg. Furthermore, since it restricts the update of local models, we discover FedProx converges slower than FedAvg from the trajectory of the global model. 
For FedCos, the global model walks around the convergence point of FedAvg so that it can achieve some points closer to the global optimum.  
For the scenes with more clients, they can be regarded as the superposition of multiple scenes with two participants, where the same phenomenon exists 
(more details in Appendix~A.1). 

Although Fig.~\ref{fig:example_com} only shows a toy example, it helps us to comprehend the advantages of FedCos as follows: \textit{FedCos can achieve the point closer to the global optimum than FedAvg}. 



At first we analyze that in each round the aggregated model of FedCos moves further in the given global direction than FedAvg's. Suppose $g(x) = 1-\frac{\langle x-\hat{x}, \hat{d}\rangle}{\|x-\hat{x} \| \cdot \| \hat{d}\|}$. The local optimization for FedCos is $f_i(x) + \mu_i g(x)$, while it for FedAvg is $f_i(x)$. 
Since rotating the coordinate system does not affect the distance and the relationship between the points or vectors, for convenience we use a matrix $X^{-1}\in \mathbb{R}^{d\times d}$ rotating the coordinate system to a new one so that $\hat{d}X = [\| \hat{d}\|, 0, \cdots, 0]^T$, i.e., the global direction is along the first dimension. For simplicity, all the analysis is in the new coordinate system, and we do not times $X$ explicitly, i.e., $x$ means $xX$. For any component $x_{(i)}$ of $x$, we have 
\small
\begin{equation}\label{eq:partialg}
	\begin{aligned}
		\frac{\partial g(x)}{\partial x_{(i)}} &= -\frac{1}{\|\hat{d}\|} \cdot \frac{d_{(i)}\| x-\hat{x}\|^2 - \langle x-\hat{x},\hat{d} \rangle (x_{(i)}-\hat{x}_{(i)})}{\|x-\hat{x}\|^3} \\
		&= -\frac{1}{\|\hat{d}\|} \cdot \frac{d_{(i)}\| x-\hat{x}\|^2 - \| \hat{d}\|(x_{(0)}-\hat{x}_{(0)}) (x_{(i)}-\hat{x}_{(i)})}{\|x-\hat{x}\|^3} \\
		&= \left\{
		\begin{aligned}
			&\frac{(x_{(0)}-\hat{x}_{(0)})^2 - \|x-\hat{x}\|^2}{\|x-\hat{x}\|^3}\leq 0  & , & i=0, \\
			&\frac{(x_{(0)}-\hat{x}_{(0)})(x_{(i)}-\hat{x}_{(i)})}{\|x-\hat{x}\|^3}  & , & i \neq 0.
		\end{aligned}
		\right.
	\end{aligned}
\end{equation}
\normalsize
From 
(\ref{eq:partialg}) the partial derivative is nonnegative in the direction of the auxiliary global direction. 

In one round that $\hat{d}^r$ is fixed, if client $i$ starts local training of FedAvg and FedCos with the same model, the local model is updated by 
(\ref{eq:iteration}) for FedAvg and 
\small
\begin{equation}\label{eq:cositeration}
	x_{i}^{r+t+1} = x_i^{r+t} - \eta_i \nabla f_i(x_i^{r+t}, B_i) - \eta_i\mu_i \nabla g(x_i^{r+t}) 
\end{equation}
\normalsize
for FedCos, where $\nabla g(x_i^{r+t}) = [\frac{\partial g(x)}{\partial x_{(0)}}, \cdots, \frac{\partial g(x)}{\partial x_{(d-1)}}]^T$. Compared to 
(\ref{eq:iteration}), equation~(\ref{eq:cositeration}) adds $-\eta_i\mu_i \nabla g(x_i^{r+t})$ at each step, which has an nonnegative component along the direction of $\hat{d}^r$~(
\ref{eq:partialg}). Intuitively speaking, all the local models of FedCos move further than FedAvg's in the direction of the global direction. The parameter $\mu_i$ controls the distance of the two points. Thus, the aggregated global model also move further than FedAvg's in this direction. 

\begin{figure}
	\centering
	\includegraphics[width = 0.95\columnwidth]{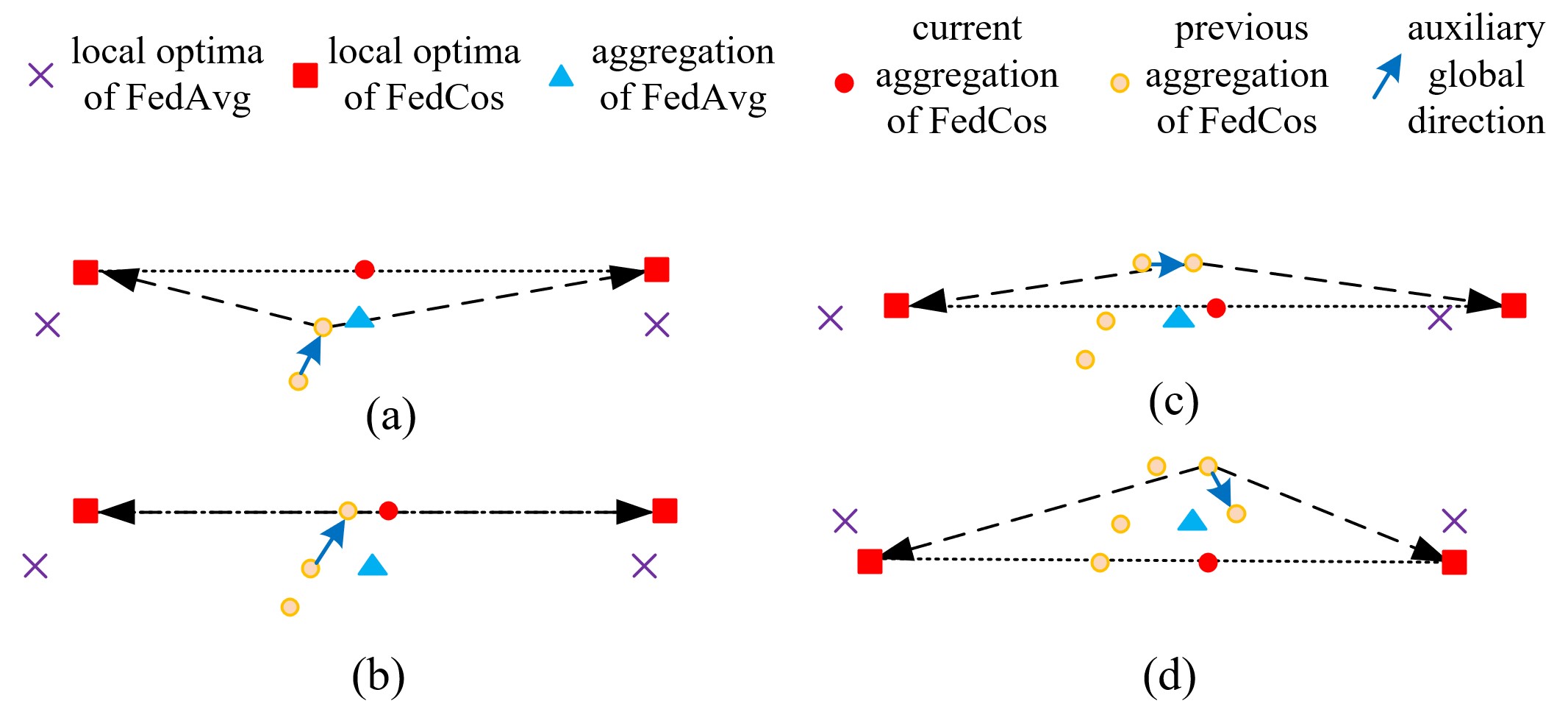}  
	\caption{The description of running FedCos.}
	\label{fig:fedcostraj} 
\end{figure}

Then we investigate the relationship of global models between FedAvg and FedCos. When FedAvg converges, i.e., the local models and the aggregated model are fixed. As shown in Fig.~\ref{fig:fedcostraj}(a), The local model of FedCos is close to the corresponding one of FedAvg, but has an offset along the global direction. The aggregated model is above the FedAvg's (named reference model). Hence in the next round, the global direction is still pointing up. Due to the constraint of $\nabla f_i(x_i^{r+t}, B_i)$, the local model of FedCos cannot move up endlessly. At some round, the local model of FedCos does not move up, and the aggregated global model does not move up too~(Fig.~\ref{fig:fedcostraj}(b)). In the next round, the global direction is no longer pointing up~(e.g., it points horizontally to the right in Fig.~\ref{fig:fedcostraj}(c)). In this case, the local model does not achieve the position as high as the previous one. The aggregated model is below the previous one~(Fig.~\ref{fig:fedcostraj}(c)). Thus in the next round, the global direction is changed to point down, and the aggregated model goes on too~(Fig.~\ref{fig:fedcostraj}(d)). In this way, the aggregated model of FedCos walks around the reference model round by round, showing the phenomenon in Fig.~\ref{fig:example_com}. The "radius" is controlled by $\mu_i$, since it is related to the distance the aggregated model moving.

From the above analysis, the global models of FedCos are around the convergence point of FedAvg. Therefore, with a proper $\mu_i$, which makes the "radius" smaller than the distance between the convergence point of FedAvg and the true global optimum, 
there are some points of FedCos closer to the true global optimum than the convergence point of FedAvg. 
In high-dimensional space, FedCos cannot iterate through all the points on the hypersphere determined by $\mu_i$, but it can explore sufficient points around the convergence point of FedAvg to explore points closer to the true global optimum with large probability.

\begin{table*}[htbp]
	\centering
		\caption{Performance (top-1 validation accuracy) comparison on CIFAR10 under different data settings, where $N=5$.} 
	\begin{tabular}{|c|c|c|c|c|}
		\hline
		& \textbf{Totally non-IID} & \textbf{90\% non-IID}   & \textbf{70\% non-IID}   & \textbf{IID}            \\ \hline
		FedAvg       & 49.56           & 64.30          & 66.82          & 68.78          \\
		FedProx(0.1) & 44.75           & 61.75          & 65.30          & 66.27          \\ 
		FedAvgM(0.5)	& 48.35		& 64.06				& 66.38		& 67.95			\\
		FedOpt(1.5)	 & 51.21		& 66.11				& 68.94		& 70.57				\\	\hline
		FedCos(0.01) & 54.54 (4.98$\uparrow$)  & 68.33 (4.03$\uparrow$) & 70.50 (3.68$\uparrow$)  & 71.75 (2.97$\uparrow$)\\
		FedCos(0.02) & 54.80 (5.24$\uparrow$)  & 69.83 (5.53$\uparrow$) & 71.46 (4.64$\uparrow$) & 73.30 (4.52$\uparrow$) \\ \hline
		FedProx(0.1)+FedCos(0.02)	& 50.85 (6.10$\uparrow$)	& 67.68 (5.93$\uparrow$)		& 70.65 (5.35$\uparrow$)		& 71.43 (5.16$\uparrow$)				\\
		FedAvgM(0.5)+FedCos(0.02)	& 54.30 (5.95$\uparrow$)	& 69.27 (5.21$\uparrow$) & 71.01 (4.63$\uparrow$)		& 72.39 (4.44$\uparrow$)					\\
		FedOpt(1.5)+FedCos(0.02)	& 54.46 (3.25$\uparrow$)	& 69.97 (3.86$\uparrow$)		& 72.64 (3.70$\uparrow$)		& 73.37 (2.80$\uparrow$)					\\		
		\hline
	\end{tabular}
	\label{table-silo-cifar}
\end{table*}

\begin{table*}[htbp]
	\centering
	\caption{Performance (top-1 validation accuracy) comparison on FMNIST under different data settings, where $N=7$.} 	
	\begin{tabular}{|c|c|c|c|c|}
		\hline
		& \textbf{Totally non-IID} & \textbf{90\% non-IID}   & \textbf{70\% non-IID}   & \textbf{IID}            \\ \hline
		FedAvg       & 74.91           & 85.15          & 86.83          & 88.04          \\
		FedProx(0.1) & 74.47           & 84.35          & 86.12          & 87.51          \\	
		FedAvgM(0.5)	& 74.76	& 85.18	& 86.65	& 87.98 \\
		FedOpt(1.5)	 &	75.47	& 85.97		& 87.60	& 88.43 \\	\hline					
		FedCos(0.01) & 79.21 (4.30$\uparrow$)  & 87.37 (2.22$\uparrow$) & 88.69 (1.86$\uparrow$)  & 89.46 (1.42$\uparrow$) \\
		FedCos(0.02) & 80.63 (5.72$\uparrow$)  & 87.61 (2.46$\uparrow$) & 89.08 (2.25$\uparrow$) & 89.52 (1.48$\uparrow$) \\ \hline		
		FedProx(0.1)+FedCos(0.02)	& 78.23 (3.76$\uparrow$)	& 87.15 (2.80$\uparrow$)	& 88.56 (2.44$\uparrow$) & 89.22 (1.71$\uparrow$) \\		
		FedAvgM(0.5)+FedCos(0.02)  & 78.72 (3.96$\uparrow$) & 87.49 (2.31$\uparrow$)	& 88.67 (2.02$\uparrow$) &89.31 (1.33$\uparrow$) \\	
		FedOpt(1.5)+FedCos(0.02)	& 80.76 (5.29$\uparrow$)	& 87.51 (1.54$\uparrow$)	& 89.03 (1.43$\uparrow$)	&89.45 (1.02$\uparrow$) \\
		\hline
	\end{tabular}
	\label{table-silo-fmnist}
\end{table*}


\section{Experiments}\label{experiments}

\subsection{Experimental settings}
\subsubsection{Baselines, Datasets and Models}
We choose FedAvg~\cite{mcmahan2017communication}, FedProx~\cite{li2018federated}, FedAvgM~\cite{hsu2019measuring, huo2020faster} and FedOpt~\cite{reddi2021adaptive} as baselines. Since we need hundreds of edge devices in our experiments, we silulate them on a server as most of previous literatures for federated learning. All methods are implemented with PyTorch1.8 and trained on GeForce RTX 3090 GPUs. We conduct experiments on 4 datasets and corresponding models as:
\begin{itemize}
	\item\textbf{CIFAR10}~\cite{krizhevsky2009learning}. CIFAR10 consists of 60000 32x32 color images in 10 categories, with 6000 images per category. The training set contains 50000 training images and the test set contains 10000 images. For CIFAR10, we use a CNN network with 2 convolutional-pooling layers plus 2 fully connected layers. In Section~\ref{furthercomparison} Lenet is used, which also shows similar results. 

	\item\textbf{FMNIST}~\cite{xiao2017fashion}. Fashion-MNIST is an updated version of MNIST. It has 28x28 grayscale images of 70000 fashion products from 10 categories. Each category has 7000 images. The training set contains 60000 images and the test set contains 10000 images. Compared to MNIST, the complexity of the task is increased. For FMNIST, we use a multi-layered perceptron network with 2 fully connected layers.

	\item\textbf{CIFAR100}~\cite{krizhevsky2009learning}. CIFAR100 consists of 100 classes, each of which has 500 images for training and 100 images for testing.
	For CIFAR100, we use ResNet18 with group normalization.
	
	\item\textbf{Shakespeare}~\cite{mcmahan2017communication} is built from The Complete Works of William Shakespeare. It consists of 143 characters with 517,106 samples. Each client takes all the samples of one character as local data. We use a two-layer LSTM classifier containing 100 hidden units with an 8D embedding layer. The task is a next-character prediction with a sequence of 80 characters as input, and there are 80 classes of characters in total.
\end{itemize}

\subsubsection{Hyperparameter setting}\label{hypersetting}
For all experiments, SGD is used as the optimizer in the local training phase, and the learning rate is 0.01 as default. 

For Table~\ref{table-silo-cifar}, 5 clients use the same hyperparameters. The batch size is 128. The number of local training steps in each round is 400. The result is after 500 epochs for each client (about 100 rounds of aggregation). For Table~\ref{table-silo-fmnist}, 7 clients are with the same settings as above. For table~\ref{table-silo-cifar100}, 5 clients are involved. The batch size is 64.  The number of local training steps in each round is 200. The result is after 300 epochs for each client. 

\begin{figure}[tbp]\small
	\centering
	\subfigure[Performance on CIFAR10]{
		\includegraphics[width=0.7\columnwidth]{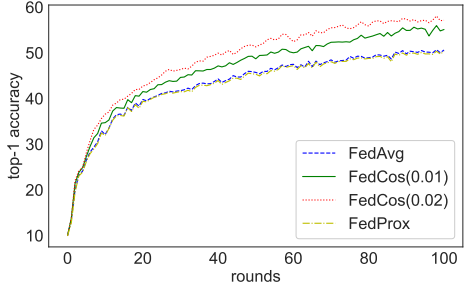}
		\label{fig:result1} 
	}
	\subfigure[Performance on FMNIST]{
		\includegraphics[width=0.7\columnwidth]{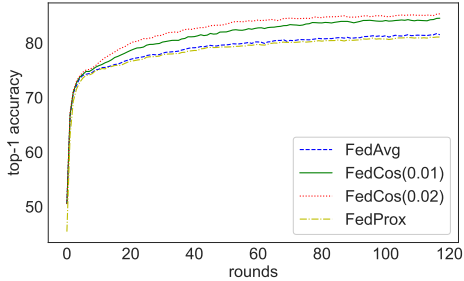} 
		\label{fig:result2} 
	}
	\caption{Performance comparison under non-IID settings where $N = 20 $.}
	\label{fig:silo_multi} 
\end{figure}

For Fig.~\ref{fig:silo_multi}, all the data are sorted by labels and divided into $2N$ equal subsets. Each client is randomly allocated 2 subsets as its local dataset. We set the batch size 64 and the number of local training steps in each round is 200. It shows the result after 500 epochs for each client.

For Table~\ref{table:cross-device} the data distribution method is the same as Section~\ref{coss-silo-multiple-clients}. We set the batch size 64 and the number of local training steps in each round is 100. All the methods run 250 rounds. For Table~\ref{table:cross-device2}, each client takes all the samples of one character as local data. The batch size is 10. The learning rate is 0.5. The number of local training steps in each round is 50.

\begin{table}[tbp]\small
	\centering
	\caption{Performance (top-1 validation accuracy) comparison on CIFAR100 under different data settings with $N=5$.} 	
	\begin{tabular}{|c|c|c|c|c|}
		\hline
		& \textbf{Totally non-IID} & \textbf{90\% non-IID} 		   	 &\textbf{IID}             \\ \hline
		FedAvg       & 50.51				&	52.84		&	64.93\\
		FedProx(0.01) & 50.34				&	50.54		&	64.58          \\ \hline
		FedCos(0.01) & \textbf{52.62}  & \textbf{54.49} & \textbf{65.13} \\
		FedCos(0.02) & \textbf{53.37}  & \textbf{55.50} & \textbf{65.35}  \\ 
		\hline
	\end{tabular}

	\label{table-silo-cifar100}
\end{table}

\begin{table*}[]\small
	\begin{minipage}{0.66\textwidth}
		\centering
		\caption{Performance comparison of CNN (the best performance of the global model in 250 rounds and the performance at the end of the training).}
		\begin{tabular}{|c|c|c|c|c|c|c|c|c|}
			\hline
			& \multicolumn{4}{|c|}{\textbf{CIFAR10}}                         & \multicolumn{4}{|c|}{\textbf{FMNIST}}                          \\ \cline{2-9} 
			& \multicolumn{2}{|c|}{\textbf{10\%}} & \multicolumn{2}{|c|}{\textbf{20\%}} & \multicolumn{2}{|c|}{\textbf{10\%}} & \multicolumn{2}{|c|}{\textbf{20\%}} \\ \cline{2-9} 
			& \textbf{Best}        & \textbf{Last}       & \textbf{Best}        & \textbf{Last}       & \textbf{Best}        & \textbf{Last}       & \textbf{Best}        & \textbf{Last}       \\ \hline
			FedAvg       & 49.91       & 48.69      & 51.44       & 50.61      & 82.69       & 79.68      & 83.29       & 79.96      \\
			FdeProx(0.1) & 49.38       & 49.63      & 50.66       & 50.38      & 82.44       & 79.73      & 83.26       & 79.68      \\ \hline
			FedCos(0.02) & \textbf{54.90}       & \textbf{52.54}      & \textbf{57.97}       & \textbf{56.15}      & \textbf{84.49}       & \textbf{82.21}      & \textbf{86.07}       & \textbf{84.35}      \\
			FedCos(0.05) & \textbf{59.36}       & \textbf{57.50}      & \textbf{61.80}       & \textbf{59.07}      & \textbf{85.69}       & \textbf{85.22}      & \textbf{86.74}       & \textbf{85.87}      \\ \hline
		\end{tabular}
		\label{table:cross-device}
	\end{minipage}
	\quad
	\begin{minipage}{0.3\textwidth}		
		\centering
		\caption{Performance comparison of two-layer LSTM after 100 rounds in cross-device FL circumstances.}		
		\begin{tabular}{|c|c|c|}
			\hline
			&\multicolumn{2}{|c|}{\textbf{Shakespeare}}  \\ \cline{2-3}
						& \textbf{10\%} & \textbf{20\%}  \\ \hline
			FedAvg       & 37.46           & 38.08                \\
			FdeProx(0.01) & 35.60           & 35.64               \\ \hline
			FedCos(0.005) & \textbf{40.25}           & \textbf{40.91}             \\
			FedCos(0.01) & \textbf{41.80}           & \textbf{42.96}              \\ \hline
		\end{tabular}
		\label{table:cross-device2}
	\end{minipage}
\end{table*}

\subsection{Performance comparison}

To compare the performance more comprehensively, we investigate the performance of the candidate methods for both FL scenarios named ``cross-silo'' and ``cross-device''~\cite{kairouz2019advances}. In cross-silo scenario, all the clients take part in the local training in each round, while in cross-device scenario only a part of clients is active in each round. We construct experimental scenes with various non-IID data settings and different number of participants to verify the effectiveness of FedCos exhaustively.

To compare the performance of FedCos, FedAvg and FedProx in different degrees of data heterogeneity, we construct 4 diverse scenes. 
(1) Totally non-IID setting: $M$ classes of data are sorted by labels and divided into $N$ equal subsets. Each client contains one as its local dataset.
(2) 90\% non-IID setting: Based on totally non-IID setting, 10\% of data of each client are extracted, shuffled and then divided into $N$ parts. Each client has one. Thus each client has 10\% of homogeneous data.
(3) 70\% non-IID setting: Like 90\% non-IID setting,  each client has 30\% of homogeneous data.
(4) IID setting: The data are randomly shuffled and divided into $N$ parts. Each client contains one.

\subsubsection{Cross-silo FL with a few number of clients}\label{coss-silo-few-clients}

Table~\ref{table-silo-cifar} shows the performance comparison of FedAvg, FedProx, and FedCos with CNN on CIFAR10, where 5 clients are involved. The penalty weight for FedProx is $0.1$.\footnote{
In these experiments, the perforance of FedProx with too small penalty weights(e.g., $0.01$) would degenerate to FedAvg.} 
For FedCos, we investigate two weights, i.e., $\mu_i=0.01$ and $0.02$. For different degrees of data heterogeneity, FedCos leads to a gain of at least 3\% over FedAvg and other 3 baselines in all cases. Then we enhance each baseline method with FedCos. It shows that FedCos can significantly improve the performance under various data heterogeneity. 
Interestingly, in the IID setting, FedCos still improves the baselines. This is because the notion of ``IID'' is based on statistics, while heterogeneity is pervasive among clients in practice. 
Table~\ref{table-silo-fmnist} illustrates the performance of MLP on FMNIST. We harvest the similar conclusions.  

We also investigate more complex medels and datasets. Table~\ref{table-silo-cifar100}\footnote{Since the performance of FedProx with penalty weight 0.1 is too poor, we use 0.01 instead.} illustrates the performance of ResNet18 (with group normalization) on CIFAR100 with $5$ clients. FedCos still significantly outperforms others.

\subsubsection{Cross-silo FL with multiple clients}\label{coss-silo-multiple-clients}
When more clients join in, FedCos still leads the way. We increase the number of clients to 20, and the results are illustrated in Fig.~\ref{fig:silo_multi}. On CIFAR10, after 500 epochs (about 100 rounds), the accuracy of FedCos with $\mu_i=0.02(0.01)$ reaches 57.14 (55.12), while the accuracy of FedAvg is only 50.63. On FMNIST, FedCos significantly improves the accuracy compared with FedAvg from 81.52 to our best result of 85.36 (84.55) with $\mu_i=0.02(0.01)$. FedProx has a similar performance to FedAvg but worse than FedCos.

\begin{figure}[tbp]\small
	\centering
	\subfigure[Reduce communication round for FedAvg]{
		\includegraphics[width=0.45\columnwidth]{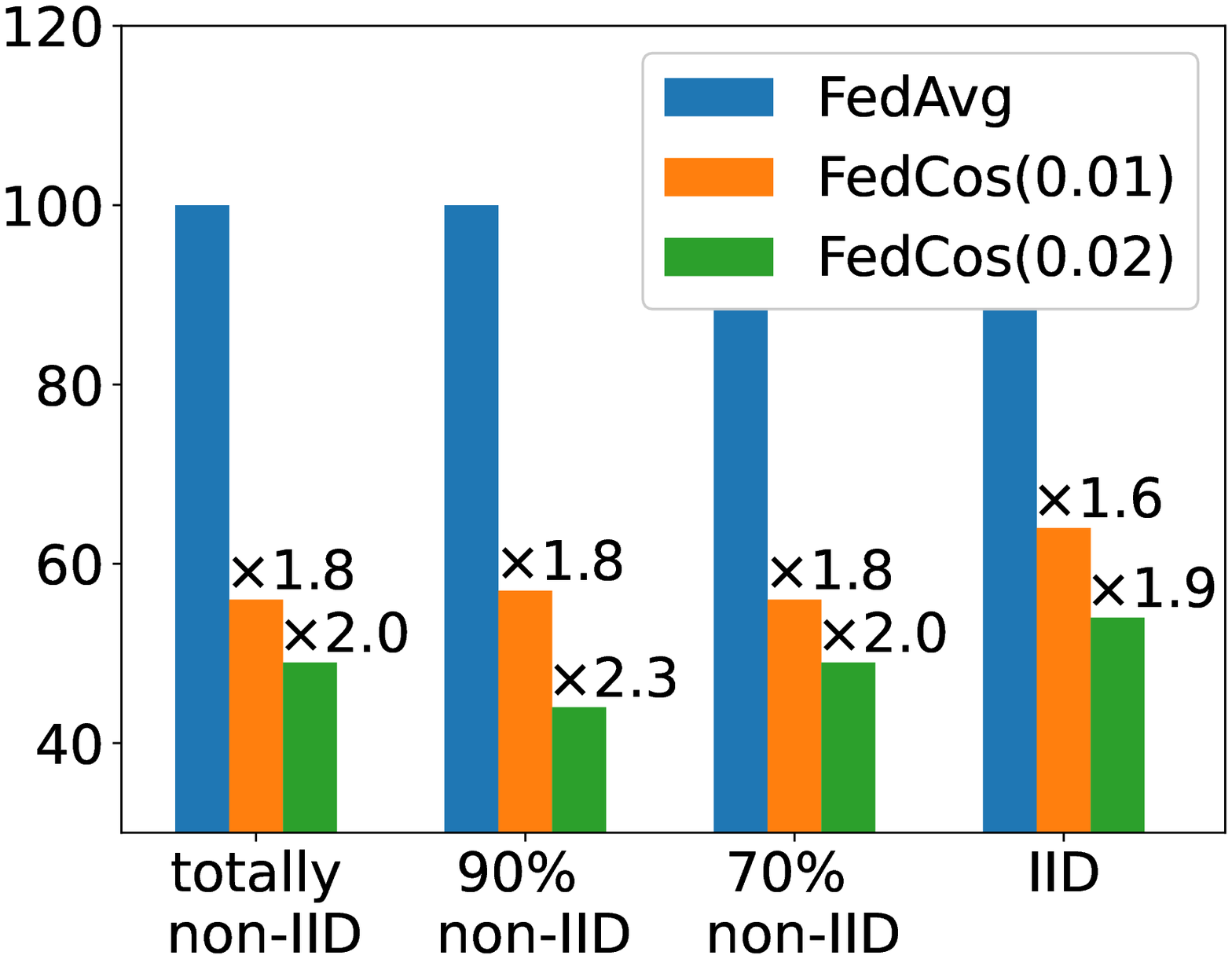}
		\label{fig:fedavg_acc} 
	}
	\subfigure[Reduce communication round for FedProx]{
		\includegraphics[width=0.45\columnwidth]{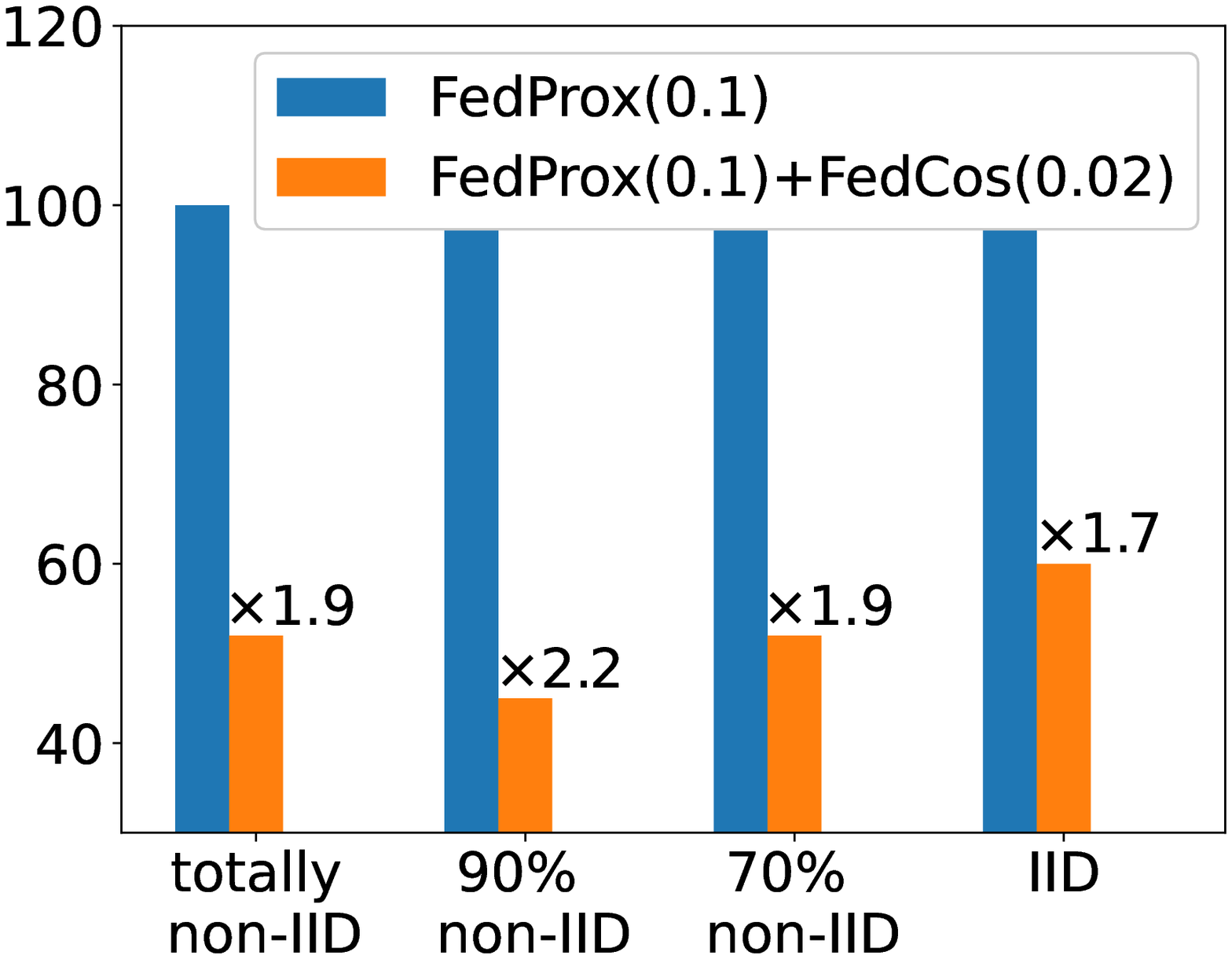} 
		\label{fig:fedprox_acc} 
	}
	\subfigure[Reduce communication round for FedAvgM]{
		\includegraphics[width=0.45\columnwidth]{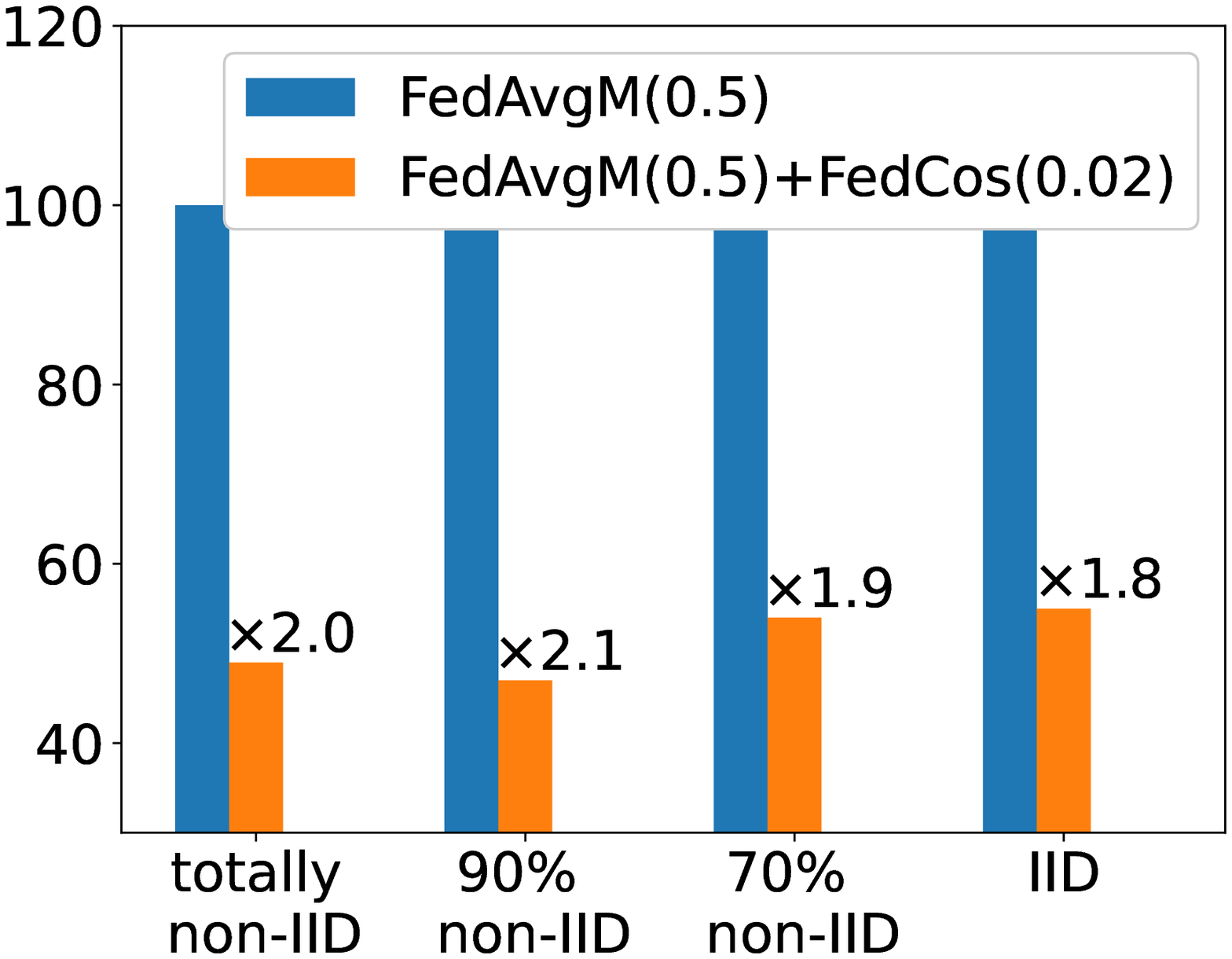} 
		\label{fig:fedavgm_acc} 
	}
	\subfigure[Reduce communication round for FedOpt]{
		\includegraphics[width=0.45\columnwidth]{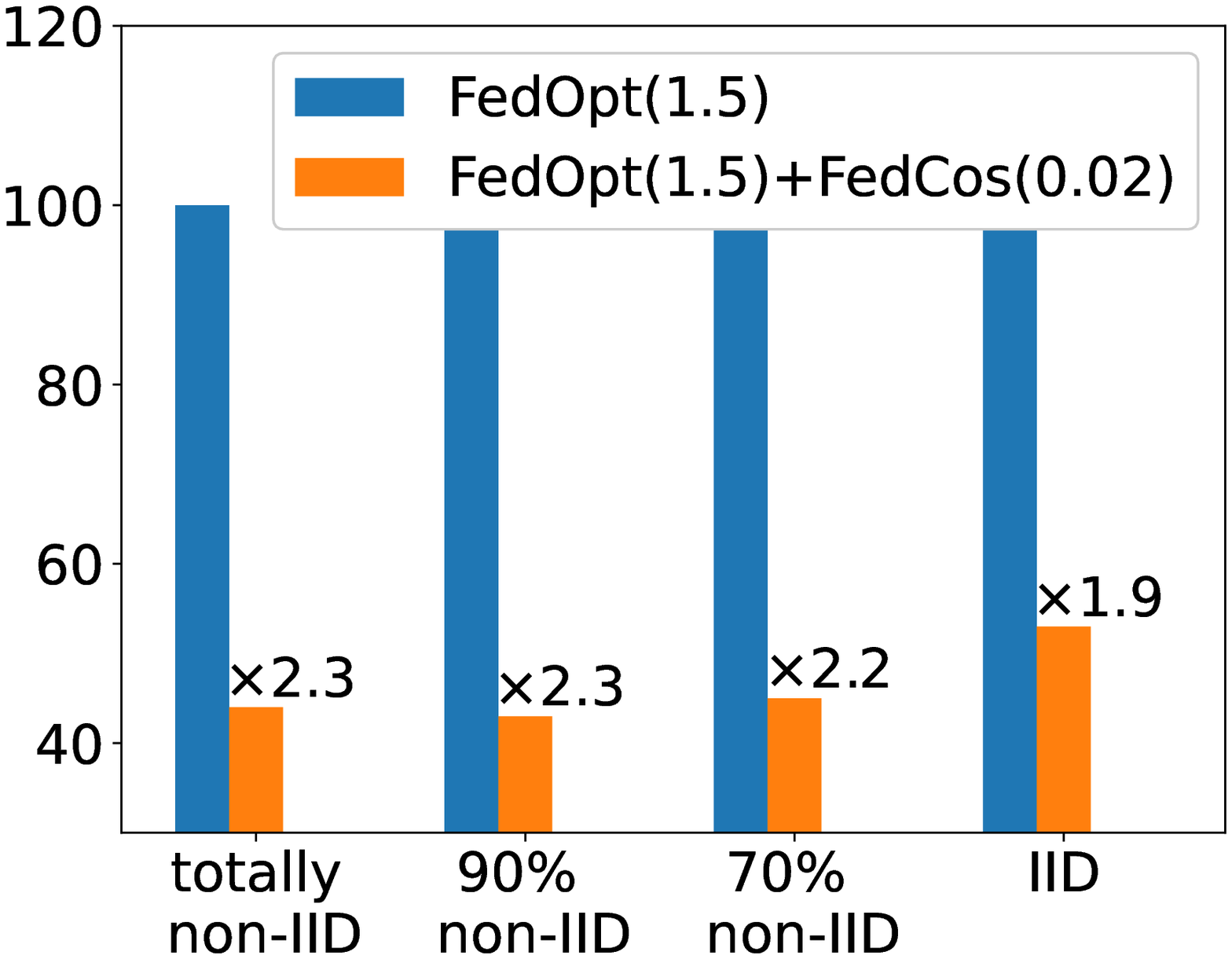} 
		\label{fig:fedopt_acc} 
	}
	\caption{FedCos enhances the communication efficiency of baselines on CIFAR10.}
	\label{fig:cifar_acc} 
\end{figure}


\begin{figure}[tbp]\small
	\centering
	\subfigure[Reduce communication round for FedAvg]{
		\includegraphics[width=0.45\columnwidth]{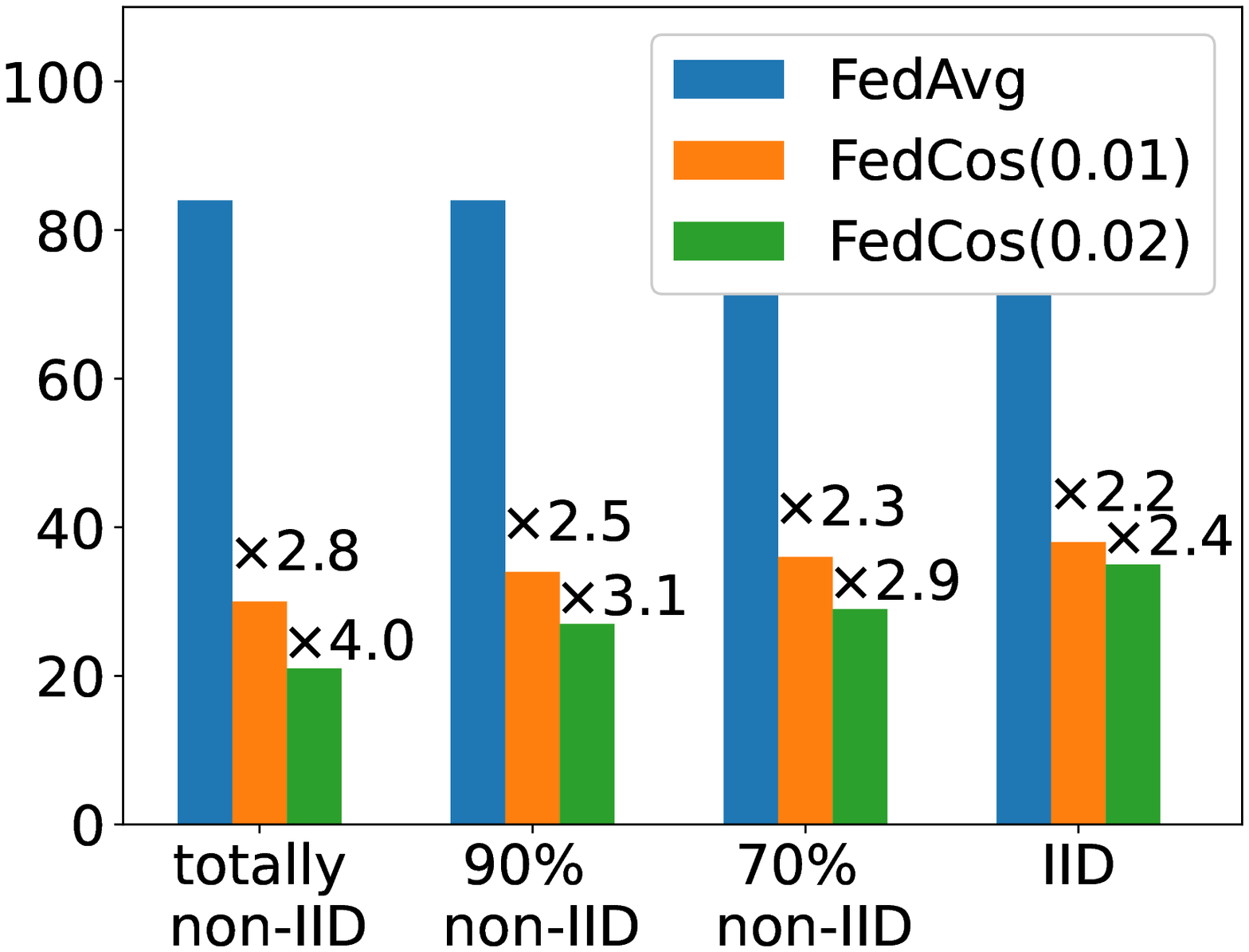}
		\label{fig:fedavg_acc2} 
	}
	\subfigure[Reduce communication round for FedProx]{
		\includegraphics[width=0.45\columnwidth]{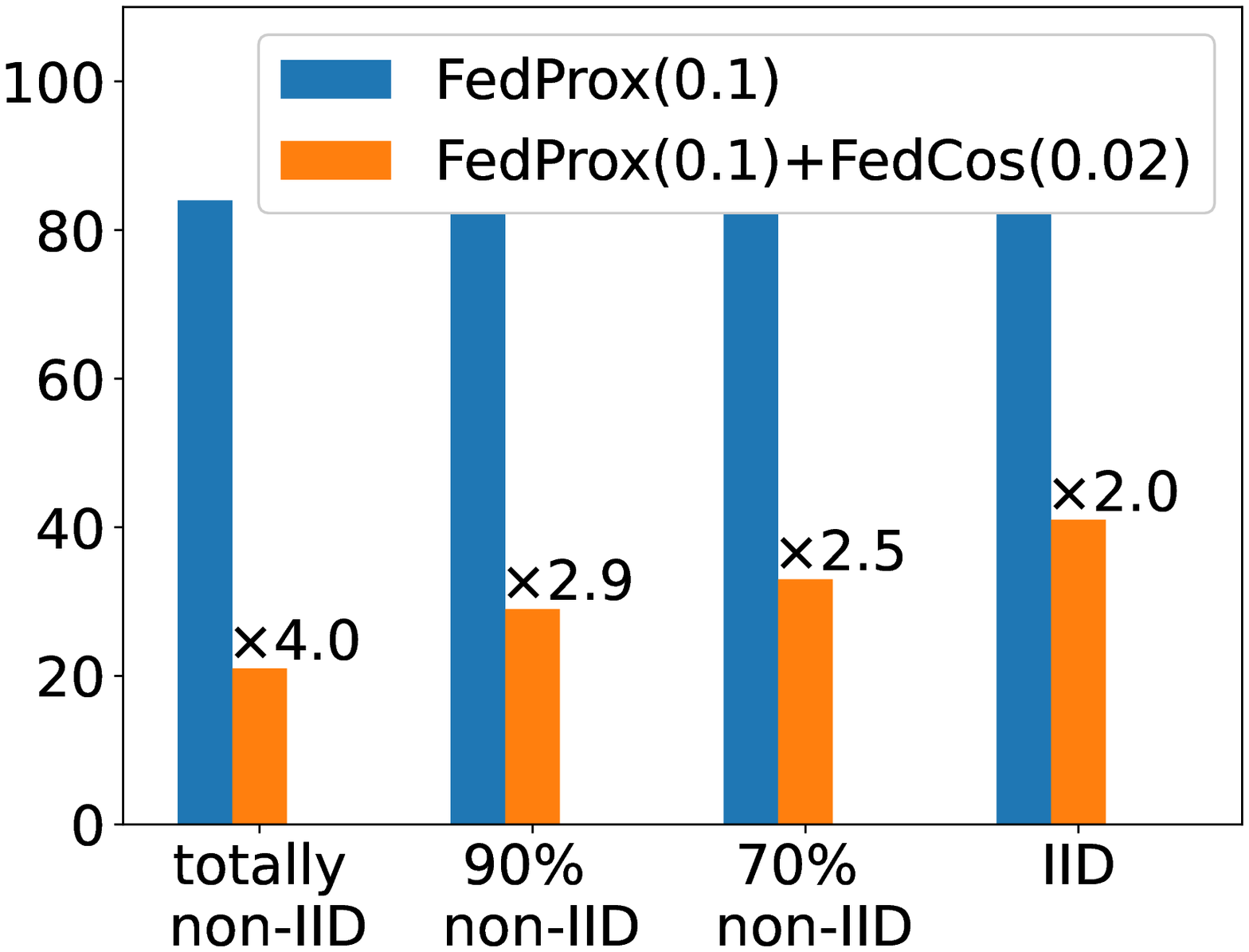} 
		\label{fig:fedprox_acc2} 
	}
	\subfigure[Reduce communication round for FedAvgM]{
		\includegraphics[width=0.45\columnwidth]{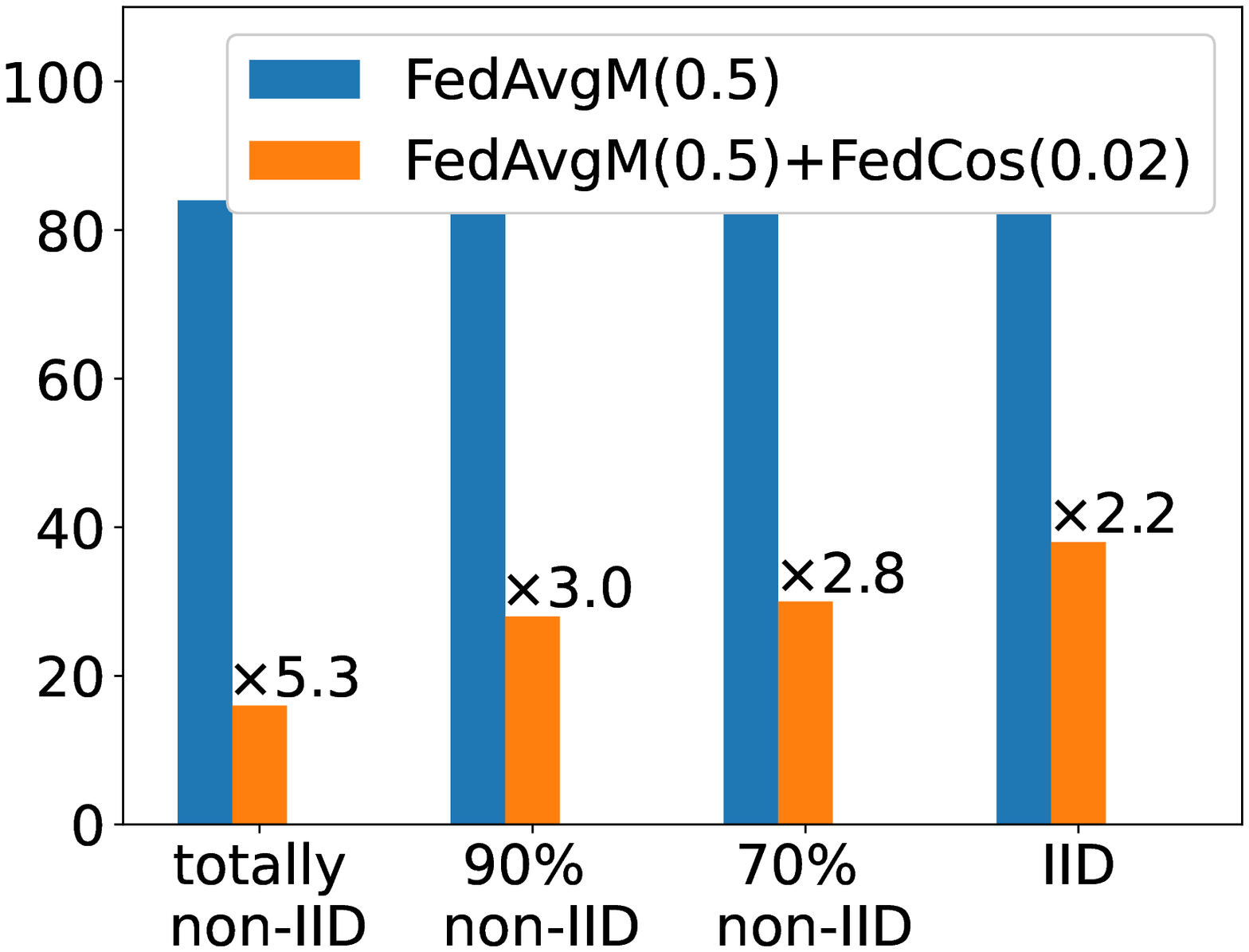} 
		\label{fig:fedavgm_acc2} 
	}
	\subfigure[Reduce communication round for FedOpt]{
		\includegraphics[width=0.45\columnwidth]{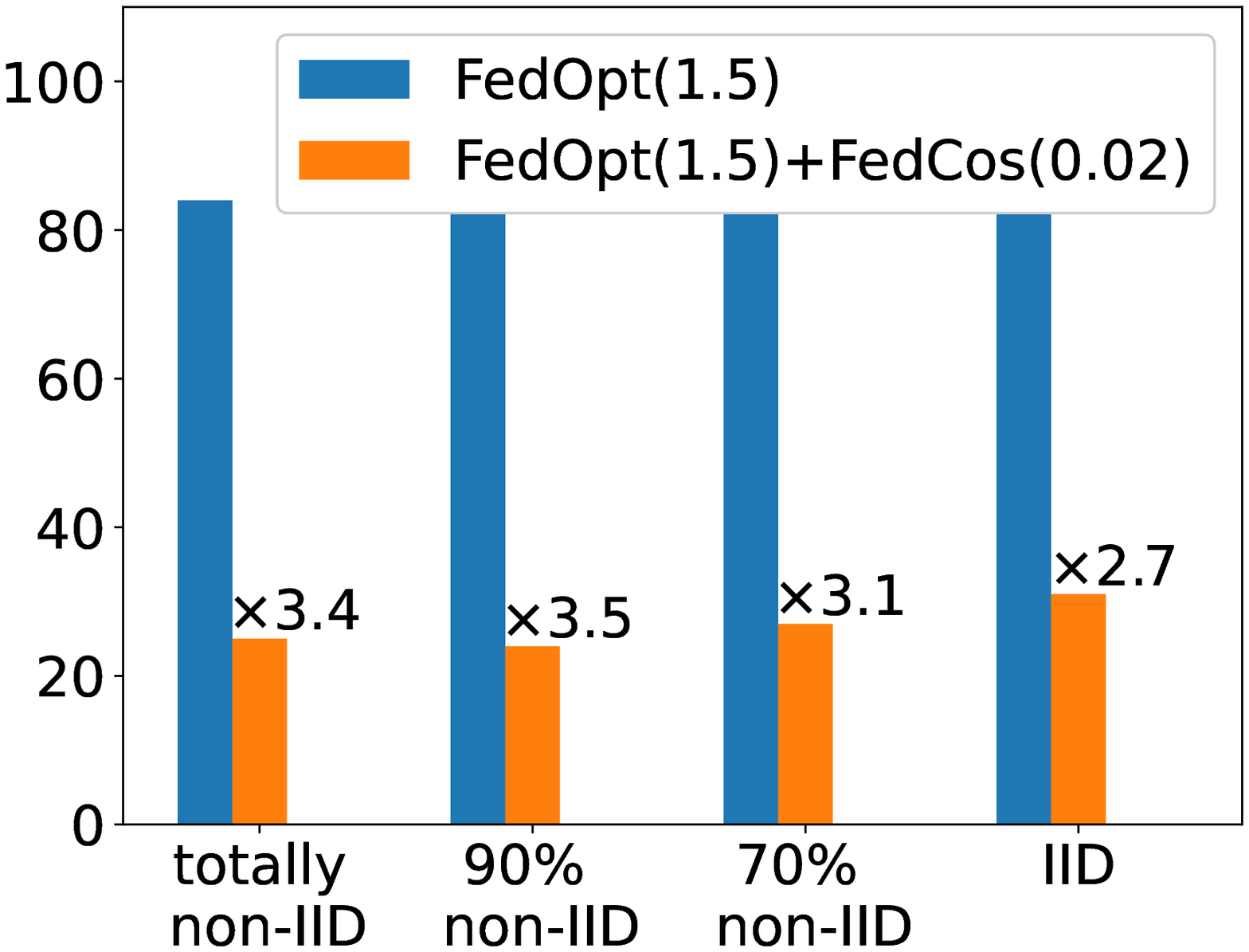} 
		\label{fig:fedopt_acc2} 
	}
	\caption{FedCos enhances the communication efficiency of baselines on FMNIST.}
	\label{fig:fmnist_acc} 
\end{figure}

\subsubsection{Cross-device FL}\label{coss-device-clients}

Table~\ref{table:cross-device} illustrates specific results, where 100 edge devices join in the FL training, and 10\%/20\% of clients are randomly chosen in each round. 
Since different clients contribute the updates in each round, the global model is not as stable as it in the cross-silo scenario. As a result, we employ a larger penalty to impose the global constraints (e.g., we set $\mu_i=0.05$). Here, we list the best results in 250 rounds and the results at the end of the training. With different proportions of participants, FedCos is superior to baselines in both measures.
Table~\ref{table:cross-device2} illustrates the performance of LSTM on Shakespeare dataset with 143 clients. 
FedCos is also better than FedAvg and FedProx.

\subsection{Comparison of communication efficiency}

FedCos can improve the communication efficiency of each rounds. With the same settings as that in Table~\ref{table-silo-cifar}, we firstly perform the 4 baselines for 100 rounds. Then we investigate how many communication rounds are required for the enhanced approaches to train the same performance models. Fig.~\ref{fig:cifar_acc} shows the results on CIFAR10, where FedCos can reduce communication rounds by 50\%. It means by the enhancement of FedCos, the communication efficiency of baselines is improved. The same results are in Fig.~\ref{fig:fmnist_acc}, where the settings on FMNIST are the same as that in Table~\ref{table-silo-fmnist}. The communication efficiency is improved by 2 to 5 times. Meanwhile, it seems that the improvement is more obvious for larger degrees of heterogeneity of data. It fits our insight that data heterogeneity leads to more serious directional inconsistency of local models, which slows down the global updates.

\subsection{Additional experiments}

\subsubsection{Detailed Comparison with FedOpt}\label{furthercomparison}

\begin{figure}[tpb]
	\centering		
	\includegraphics[width=0.7\columnwidth]{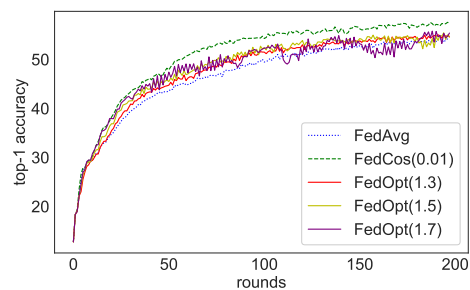}
	\caption{Comparison of FedCos and FedOpt.}
	\label{fig:fedlen} 
\end{figure}

\begin{figure}[tbp]
	\centering
	\includegraphics[width = 0.7\columnwidth]{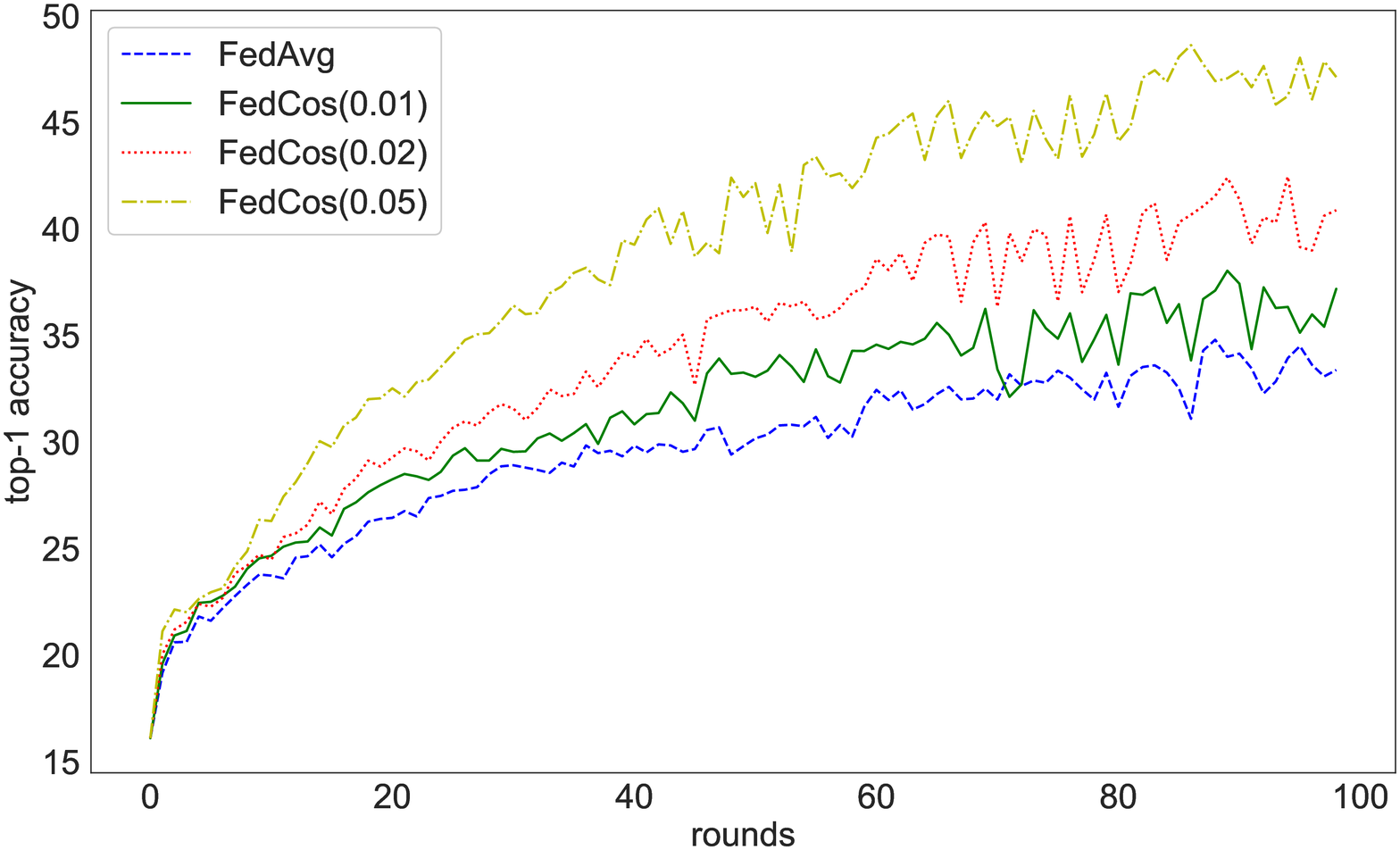}  
	\caption{Performance comparison for ResNet18.}
	\label{fig:comparison_resnet18} 
\end{figure}

\begin{figure}[tbp]
	\centering
	\includegraphics[width = 0.7\columnwidth]{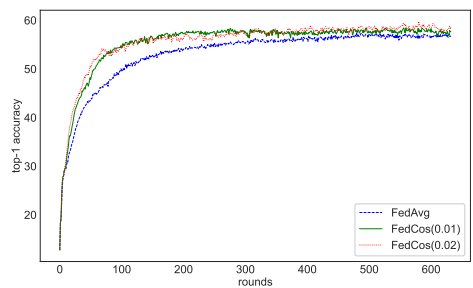}  
	\caption{Performance comparison with 3200 epoches.}
	\label{fig:comparison_long} 
\end{figure}

\begin{figure*}[tbp]
	\centering
	\subfigure[Moving distance of one local model in local training]{
		\includegraphics[width=0.62\columnwidth]{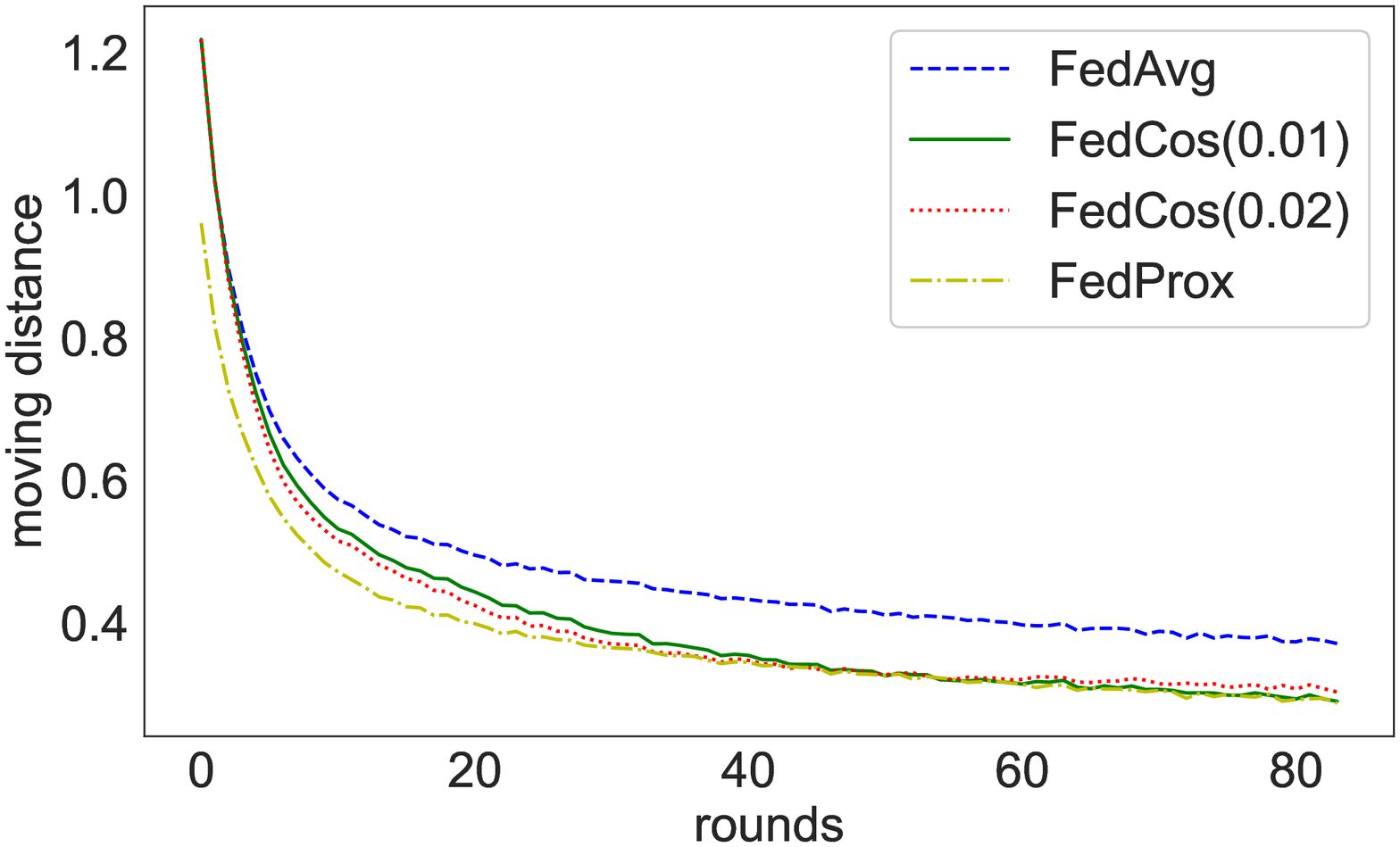}
		\label{fig:fedcospro2} 
	}
	\subfigure[The distance between two local models]{
		\includegraphics[width=0.62\columnwidth]{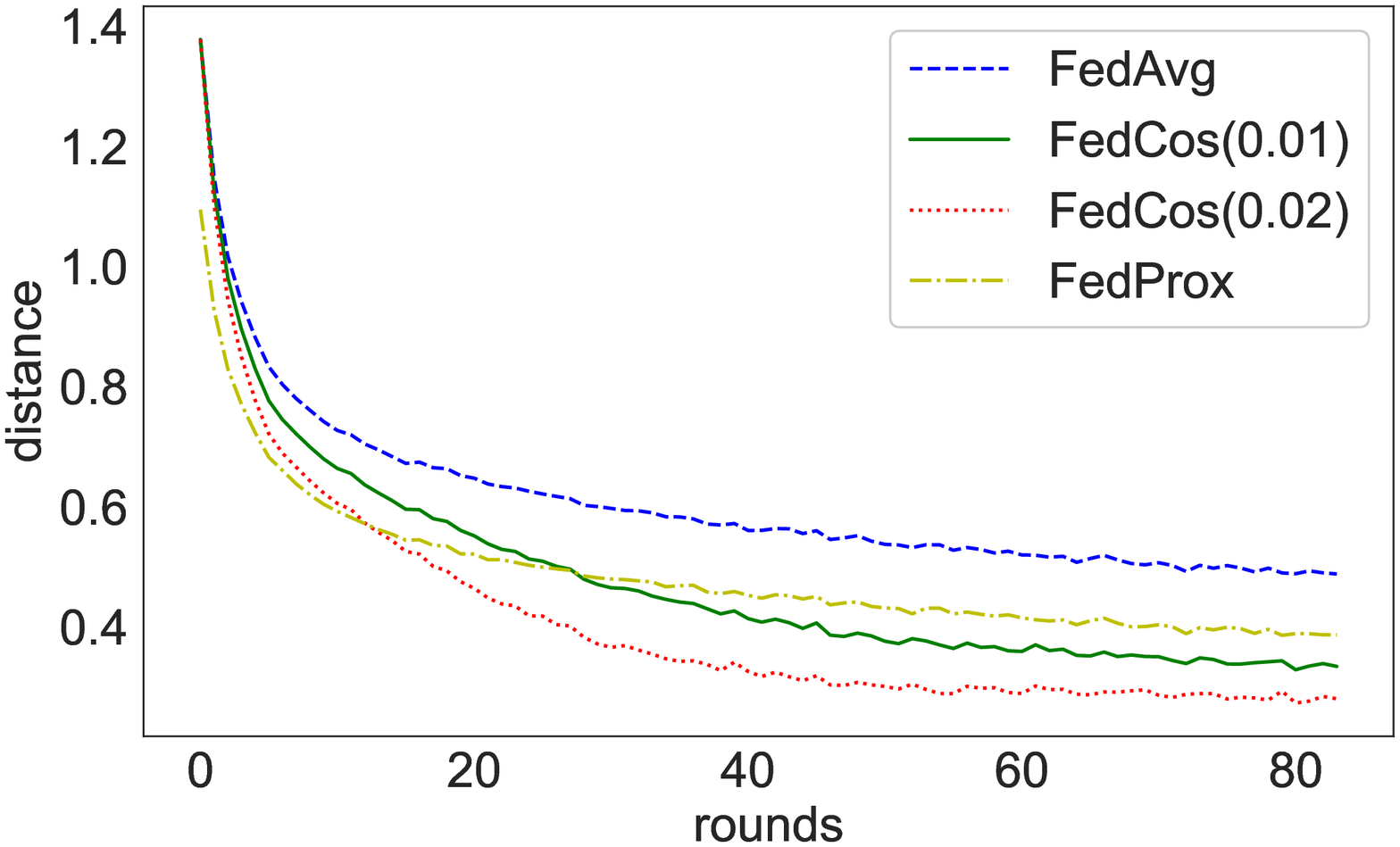} 
		\label{fig:fedcospro4} 
	}
	\subfigure[Moving distance of the global model]{
		\includegraphics[width=0.62\columnwidth]{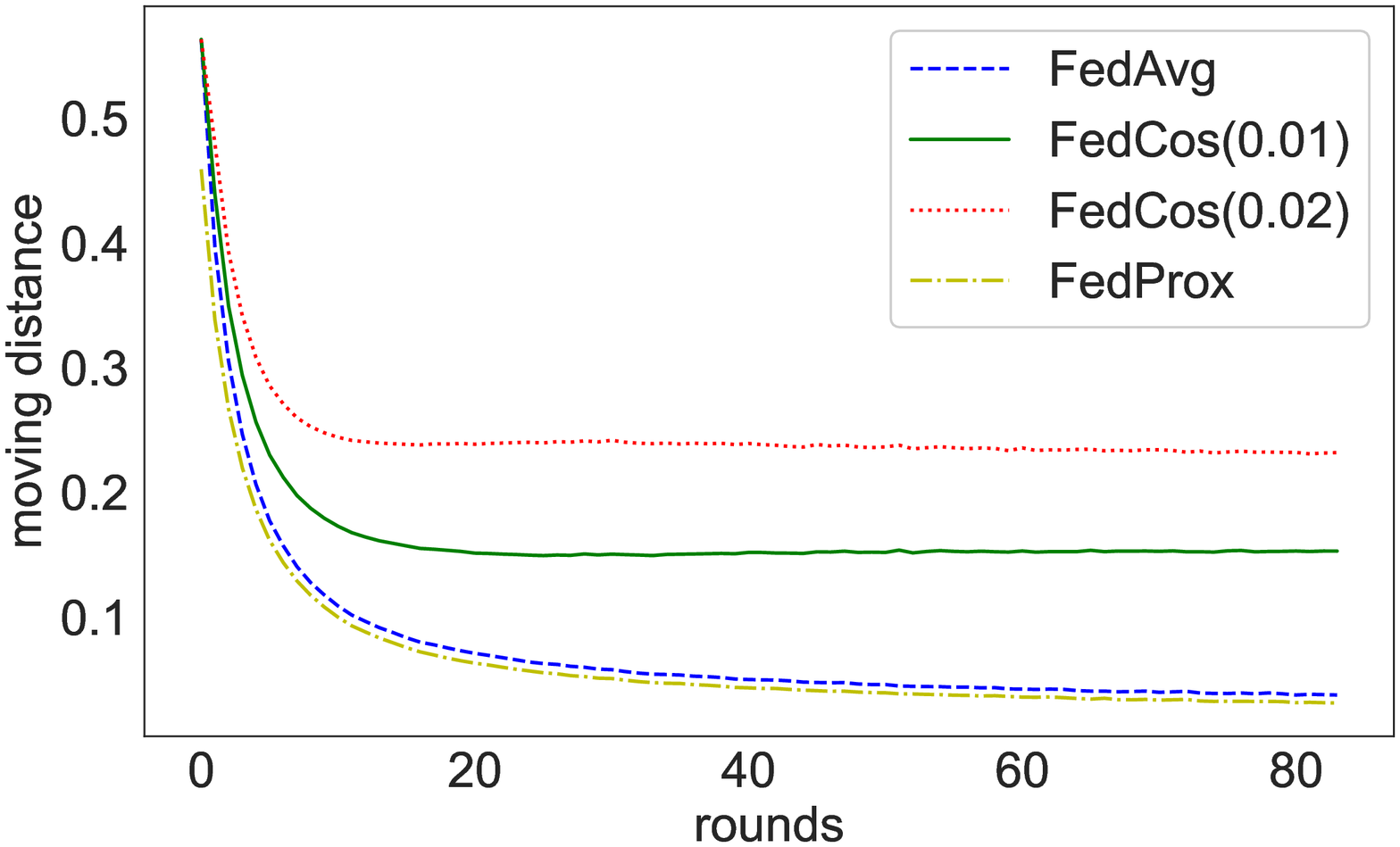} 
		\label{fig:fedcospro1} 
	}
	\caption{Mechanism illustration of FedCos.}
	\label{fig:fedcospro} 
\end{figure*}

From Table~\ref{table-silo-cifar} and Table~\ref{table-silo-fmnist}, FedOpt performs best in 4 baselines, and it can also increase the moving distance of the global model with carefully selected hyperparameters. FedOpt adjusts the learning rate of global updating $\eta_g$, where FedAvg is the case with $\eta_g=1$. Fig.~\ref{fig:fedlen} shows the performance comparison between FedCos and FedOpt with various $\eta_g$. For the first several rounds, FedOpt speeds up the training like FedCos. However, no matter what the learning rate is, FedOpt performs as same as FedAvg after 200 rounds. In contrast, FedCos keeps ahead of FedAvg and FedOpt all the time. Furthermore, for FedOpt a larger $\eta_g$ would make the training instability. In our experiment, when $\eta_g = 1.7$, it has apparent jitter on the performance curve. As $\eta_g$ continues to increase, the performance would seriously decrease.  


\subsubsection{Performance comparison under deeper neural networks}\label{deepnn}

To demonstrate the applicability of FedCos, the performance for more deeper neural networks is investigated. Fig.\ref{fig:comparison_resnet18} shows the results under total non-IID setting on CIFAR10 for ResNet18, which is typical deep neural network much more complex than previously evaluated models. Same as before, FedCos with various of panelty weights outperforms FedAvg much more. It should be noted that, since batch normalization does not suit the Non-IID setting, we replace it with alternative group normalization mechanism~\cite{hsieh2020non}. Due to the high computational overhead, most of previous works explore the mechanism of FL by simple networks.

\subsubsection{Performance comparison under long time training}\label{longtime}

In order to investigate the influence of round number, we greatly increase the training time. Fig.~\ref{fig:comparison_long} shows the comparison under totally non-IID setting on CIFAR10, where each client iterates 3200 epochs (more than 600 rounds). At the end of the training, FedCos with $\mu_i=0.02(0.01)$ achieves 58.58(57.66), while FedAvg is only 56.83. For the best result on the test dataset, FedCos with $\mu_i=0.02(0.01)$ achieves $59.73(58.48)$, while FedAvg is only 57.45. Although the gap between FedCos and FedAvg is narrowed, FedCos still outperforms FedAvg at the end of training. In practice, it is no need to spend so much computational overhead and communication cost, which can be saved by fewer local iteration steps and aggregation round. So it does not affect the results of the previous comparison.

\subsection{Mechanism visualization}
As noted in the analysis of previous section, 
the directional inconsistency of local models is the crucial factor degrading the performance of FL methods in non-IID scenarios, and FedCos eases the inconsistency by the given global direction vector.
Fig.~\ref{fig:consistency} and Fig.~\ref{fig:fedcospro} further verify the results under the totally non-IID scenario in Table~\ref{table-silo-fmnist}. Specifically, Fig.~\ref{fig:fedcospro2} shows the moving distance of one local model. For FedCos, the moving distance of the local model is smaller than it for FedAvg, and the angle between the moving directions of two local models is smaller (Fig.~\ref{fig:consine_noniid3}). 
Therefore, the distance of any two local models for FedCos is smaller than FedAvg's (Fig.~\ref{fig:fedcospro4}), \textit{which validates the analysis in Fig.~\ref{fig:fl_move}(c)}. 
FedCos obtains the local optima (approximately) closer to each other than FedAvg. Thus, the aggregated model is closer to all the local optima and performs better performance.


The global model only has little changes for FedAvg in each round, while it moves farther for FedCos, as shown in Fig.~\ref{fig:fedcospro1}. Thus the global model updates of FedCos are more efficient, \textit{which confirms the view in Fig.~\ref{fig:fl_move}(b)}. 
Moreover, although the moving distance of the local model for FedProx is small (Fig.~\ref{fig:fedcospro2}), the angle is large and comparable with FedAvg's (Fig.~\ref{fig:consine_noniid3}). Accordingly, the distance between the two local models is not as small as expected (Fig.~\ref{fig:fedcospro4}). 
Meanwhile, because of the strong constraint, the local model for FedProx degrades the efficiency of local training. 
FedProx performs worse than FedCos/FedAvg, despite with smaller local models' distances than FedAvg's.

\section{Conclusion}\label{conclusion}

In this work, we propose FedCos to improve model accuracy in FL. FedCos introduces an auxiliary global direction to guide the dirction of local model in training. This scheme improves the performance of the aggregated model by reducing the directional inconsistency of local models. We analyze the properties of FedCos, and point out that FedCos can obtain better models than the standard FL method FedAvg. From the experiments, FedCos vastly outperforms FedAvg in a variety of FL scenes. Due to the complexity of analysis, in this paper more rigorous theoretical proofs are not given, which would be explored in future work.


%





\bibliography{fedcos}
\bibliographystyle{IEEEtran}

\section{Appendix}

\subsection{FL example with multiple clients}\label{example_multiclient} 

\begin{figure}[htp]
	\centering
	\includegraphics[width = 0.9\columnwidth]{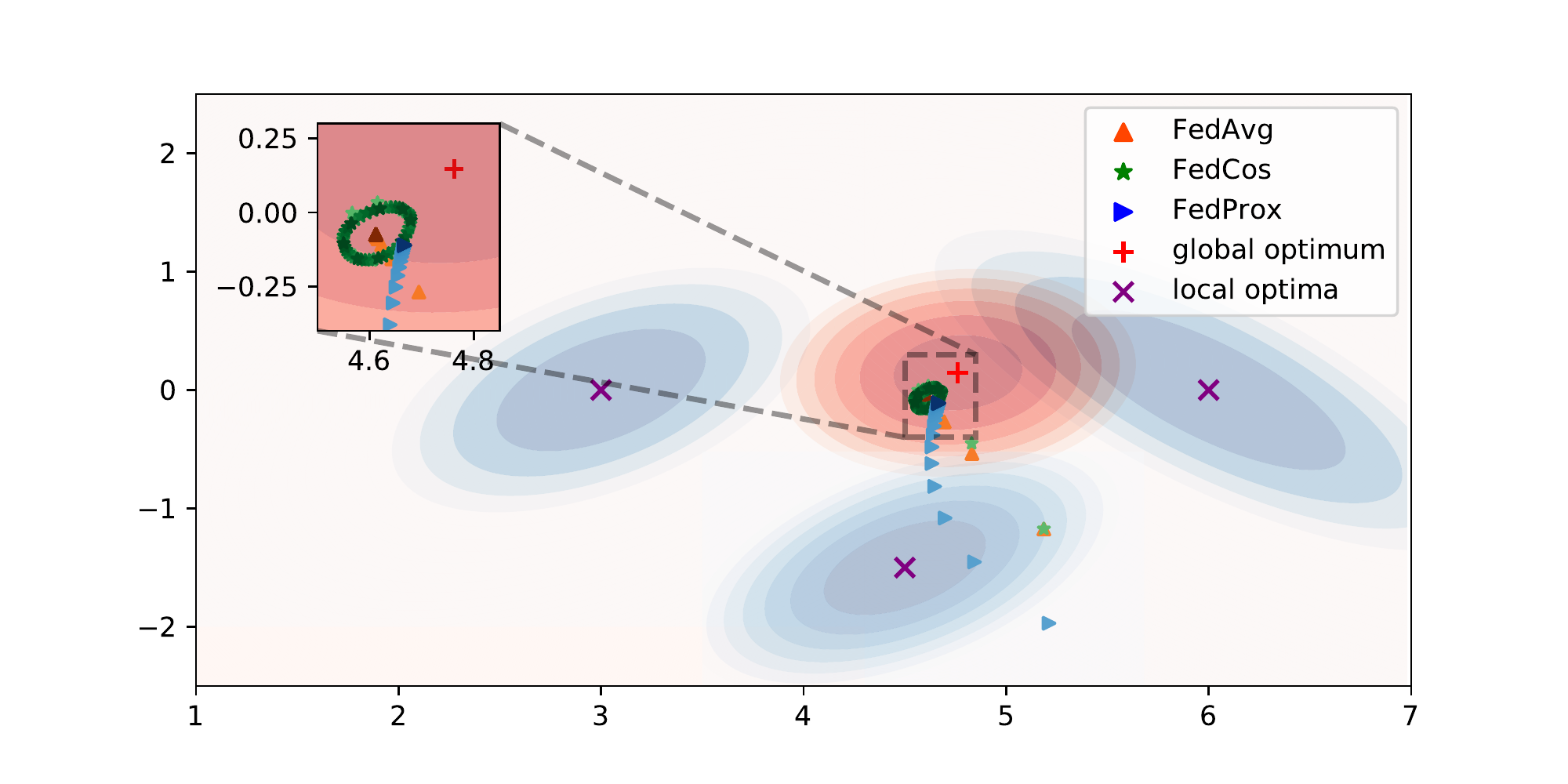}  
	\caption{Comparison of the trajectories of global models by FedAvg, FedProx and FedCos on a simple example with 3 participants. All the methods perform 80 rounds. The points with darker colors denote the models from later rounds.}
	\label{fig:example_com_3clients} 
\end{figure}

For the simple federated example with more participants, the same phenomenon exists. Fig.~\ref{fig:example_com_3clients} illustrates 3 clients. The initial point is (4.53, 0.38). As in the example in Fig.~\ref{fig:example_com}, the trajectory of the global model in FedCos is around the stationary point of FedAvg. FedCos can obtain better models closer to the global optimum than FedAvg.

\end{document}